\PassOptionsToPackage{unicode}{hyperref}
\PassOptionsToPackage{hyphens}{url}
\documentclass[
  11pt,
]{article}
\usepackage{xcolor}
\usepackage[margin=1in]{geometry}
\usepackage{amsmath,amssymb}
\usepackage{graphicx} 
\usepackage{float} 
\usepackage{fancyhdr} 
\usepackage{ragged2e}

\usepackage{iftex}
\ifPDFTeX
  \usepackage[T1]{fontenc}
  \usepackage[utf8]{inputenc}
  \usepackage{textcomp} 
\else 
  \usepackage{unicode-math} 
  \defaultfontfeatures{Scale=MatchLowercase}
  \defaultfontfeatures[\rmfamily]{Ligatures=TeX,Scale=1}
\fi
\usepackage{lmodern}
\ifPDFTeX\else
\fi
\IfFileExists{upquote.sty}{\usepackage{upquote}}{}
\IfFileExists{microtype.sty}{
  \usepackage[]{microtype}
  \UseMicrotypeSet[protrusion]{basicmath} 
}{}
\makeatletter
\@ifundefined{KOMAClassName}{
  \IfFileExists{parskip.sty}{%
    \usepackage{parskip}
  }{
    \setlength{\parindent}{0pt}
    \setlength{\parskip}{6pt plus 2pt minus 1pt}}
}{
  \KOMAoptions{parskip=half}}
\makeatother
\usepackage{longtable,booktabs,array}
\usepackage{calc} 
\usepackage{etoolbox}
\makeatletter
\patchcmd\longtable{\par}{\if@noskipsec\mbox{}\fi\par}{}{}
\makeatother
\IfFileExists{footnotehyper.sty}{\usepackage{footnotehyper}}{\usepackage{footnote}}
\makesavenoteenv{longtable}
\providecommand{\tightlist}{%
  \setlength{\itemsep}{0pt}\setlength{\parskip}{0pt}}
\usepackage{bookmark}
\IfFileExists{xurl.sty}{\usepackage{xurl}}{} 
\hypersetup{
  hidelinks,
  pdfcreator={LaTeX via pandoc}}

\author{}
\date{}

\begin{document}

\section{The Topology of Multimodal Fusion: Why Current Architectures
Fail at Creative
Cognition}\label{the-topology-of-multimodal-fusion-why-current-architectures-fail-at-creative-cognition}

\textbf{Xiujiang Tan}

Guangzhou Academy of Fine Arts

\textbf{Preprint v7 · April 2026}

\begin{center}\rule{0.5\linewidth}{0.5pt}\end{center}

\begin{center}\textbf{\large Abstract}\end{center}\label{abstract}

This paper identifies a structural limitation in current multimodal AI
architectures that is topological rather than parametric in nature.
Contrastive alignment (CLIP paradigm), cross-attention fusion
(GPT-4V/Gemini paradigm), and diffusion-based generation share a common
geometric prior: \emph{modal separability}---the assumption that the
relationship between modalities is an interface relation rather than a
constitutive one. We term this representational regime \emph{contact
topology}.

Our argument rests on three mutually reinforcing theoretical pillars.
\textbf{Philosophical pillar} (the generative center): Wittgenstein's
saying/showing distinction identifies the structural boundary of
propositional representation, and the Chinese craft epistemology
tradition centered on \emph{xiàng} (operative schema)---reinterpreted
here not as a phenomenological correlate but as the \emph{third state}
emerging at the saying/showing intersection---provides the cruciform
framework (\emph{dào}/\emph{qì} × saying/showing) from which the
mathematical and computational pillars derive as natural instantiations.
\textbf{Cognitive science pillar}: empirical findings on the tripartite
co-activation of the Default Mode Network (DMN), Executive Control
Network (ECN), and Salience Network (SN), with the SN reinterpreted as a
coupling regulator rather than a simple network switch.
\textbf{Mathematical pillar}: connection theory on fiber bundles over
the space of modality configurations, where connection curvature
quantifies cross-domain isomorphism and the Yang-Mills action functional
demarcates three dynamical regimes.

By introducing the topological opposition between overlap isomorphism
(transversal coupling of propositional/non-propositional dimensions in
creative emergence) and superimposition collapse (loss of transversality
in cognitive disorganization), the theory acquires a precise
falsification condition and a pathological mirror. We advance the
\emph{Overlap Zone Isomorphism Conjecture} in a three-level nested
hierarchy and design a four-experiment cognitive neuroscience testing
framework employing a dimensional individual-differences design. On the
computational side, we provide a UOO implementation path based on Neural
ODEs with persistent homology topological regularization---including a
scalability roadmap from direct Vietoris-Rips computation to
landmark-based Witness Complexes and Distance-to-Measure filtration---a
synthetic-data minimum viable proof of concept executable within weeks,
the ANALOGY-MM cross-modal analogy benchmark with a novel
error-type-ratio metric and a scalable forced-choice evaluation
protocol, and predict UOO-specific failure modes under three stress
conditions. The experimental methodology incorporates a dual-track
causal analysis (Dynamic Causal Modeling on fMRI, Granger Causality on
MEG) to test the directionality of SN gating before committing to
invasive TMS intervention. A phased experimental roadmap with explicit
termination criteria at each gate ensures the program exits cleanly if
falsified.

\textbf{Keywords}: multimodal AI; fiber bundles; gauge equivariance;
non-separable representations; topological data analysis; persistent
homology; cognitive neuroscience of creativity; cross-modal transfer;
pathological mirror; Yang-Mills action functional; structural tension
index; ANALOGY-MM; Dynamic Causal Modeling; Granger causality; Witness
Complex; cruciform structure; xiàng; chuànghuà; huàcái; META-TOP

\begin{center}\rule{0.5\linewidth}{0.5pt}\end{center}

\subsection{1. Introduction: The Architectural
Ceiling}\label{introduction-the-architectural-ceiling}

\subsubsection{1.1 Delimiting the Problem
Domain}\label{delimiting-the-problem-domain}

The capability ceiling of current AI systems is most pronounced in
domains where meaning does not reside at the level of human-language
description: molecular dynamics, high-dimensional sensor fusion,
spatiotemporal gene expression atlases, and creative cross-modal
reasoning. These domains share a structural characteristic: their
informational content demands \emph{simultaneous} processing of
dimensions that current architectures treat as separable channels.

The nature boundary of this limitation must be stated precisely. The
domains listed above also lack large-scale labeled datasets and
well-established benchmarks, and involve complex physical
constraints---these are engineering bottlenecks superimposed on the
architectural problem. Protein structure prediction was once regarded as
an insurmountable barrier, until AlphaFold2 conquered it with a
domain-specific architecture and a fully separable representational
scheme. Our precise claim is therefore not that current systems are
constitutively incapable of handling these domains. Rather: for a
specific class of cross-domain reasoning tasks---those requiring
essentially non-separable representations---current architectures'
inductive biases and training objectives \textbf{systematically
suppress} the representational structure required for optimal
performance. This more precisely worded claim is also more strongly
falsifiable.

This distinction---between constitutive impossibility and systematic
suppression---is strategically critical. It implies that the
computational path forward may not require an entirely new architecture,
but rather new training objectives and regularization schemes
specifically designed to \emph{maintain and reinforce} non-separable
representations. The topological regularization proposed in §6 is a
direct response to this insight.

\subsubsection{1.2 Architectural Diagnosis: Three Strategies and a
Shared
Prior}\label{architectural-diagnosis-three-strategies-and-a-shared-prior}

Current multimodal AI employs three principal integration strategies.
Following the principle established by the Geometric Deep Learning (GDL)
framework---that the deepest architectural choice in deep learning is
the symmetry group the system presupposes for its processing domain
(Bronstein et al., 2021)---we analyze the geometric prior encoded by
each strategy.

\textbf{Strategy 1: Contrastive alignment (CLIP paradigm).} Vision and
language are projected into independent embedding spaces and aligned via
cosine similarity maximization. The geometric prior is explicit: each
modality resides on an independent manifold connectable by linear
projection. The global minimum of the cosine similarity loss corresponds
precisely to fully separable embedding pairs---this is contact topology
in its most literal form: two manifolds meeting at zero-dimensional
contact points.

\textbf{Strategy 2: Cross-attention fusion (GPT-4V/Gemini paradigm).}
One modality sequence queries another through key-value attention.
Admittedly, bidirectional cross-attention and self-attention applied to
concatenated sequences---as in the Gemini architecture---have eliminated
the query/key-value asymmetry of earlier systems. Vision and language
tokens are processed in a single unified sequence; the predefined
``questioner'' and ``answerer'' roles no longer exist.

However, a subtler form of separability persists: even under symmetric
attention, representations maintain \textbf{modality
traceability}---each token can always be traced back to its modal
origin. This is more covert than query/key-value asymmetry yet harder to
eliminate by simple architectural modification. It constitutes the
residual form of contact topology in symmetric attention architectures.
More formally, in any layer where token embeddings are concatenated from
distinct encoders, the token index itself carries modal provenance
information, and the attention softmax operates on this
provenance-tagged representation at every step. The separability is thus
encoded not in the attention mechanism per se, but in the
\emph{indexical structure} of the input representation.

\textbf{Strategy 3: Diffusion-based generation.} Diffusion models
traverse highly entangled latent states during the denoising process.
The crucial point: the architecture itself does not preclude dwelling in
the entangled state---what drives the system out is the denoising
objective that demands a clearly interpretable output image. With a
modified training objective---for instance, introducing representation
quality metrics at intermediate denoising steps---diffusion
architectures can be trained to \emph{dwell} in entangled states. This
observation is key: the limitation is in the loss function, not the
representational capacity.

\textbf{Shared prior: Modal separability.} All three strategies share
the same geometric prior: the inter-modal relationship is an
\emph{interface relation} (contact topology) rather than a
\emph{constitutive relation} (overlap topology). This prior
systematically suppresses overlap-zone representations at the level of
training objectives and inductive biases, rather than constitutively
excluding them at the architectural level.

\subsubsection{1.3 Core Analogy: Contact Topology vs.~Overlap
Topology}\label{core-analogy-contact-topology-vs.-overlap-topology}

We crystallize the diagnosis through a topological distinction:

\begin{itemize}
\tightlist
\item
  \textbf{Contact topology}: Two manifolds meet at a boundary of measure
  zero. Information can be \emph{exchanged} across the interface but can
  never be \emph{co-constituted} in an interior region.
\item
  \textbf{Overlap topology}: A representational region with non-zero
  interior (Lebesgue measure \textgreater{} 0) in which elements from
  two modalities are structurally inseparable, mutually constituting
  each other's semantic content.
\end{itemize}

Current architectures operate exclusively via contact mechanisms. The
Overlap Zone hypothesis asserts that an important class of cognitive and
computational operations requires the overlap mechanism. The remainder
of the paper develops this hypothesis through a philosophical foundation
that serves as the generative center (§3), with neuroscientific (§4),
mathematical (§5), computational (§6--7), and experimental (§8) pillars
deriving from it as natural instantiations.

\begin{figure}[htbp]
\centering
\includegraphics[width=\textwidth]{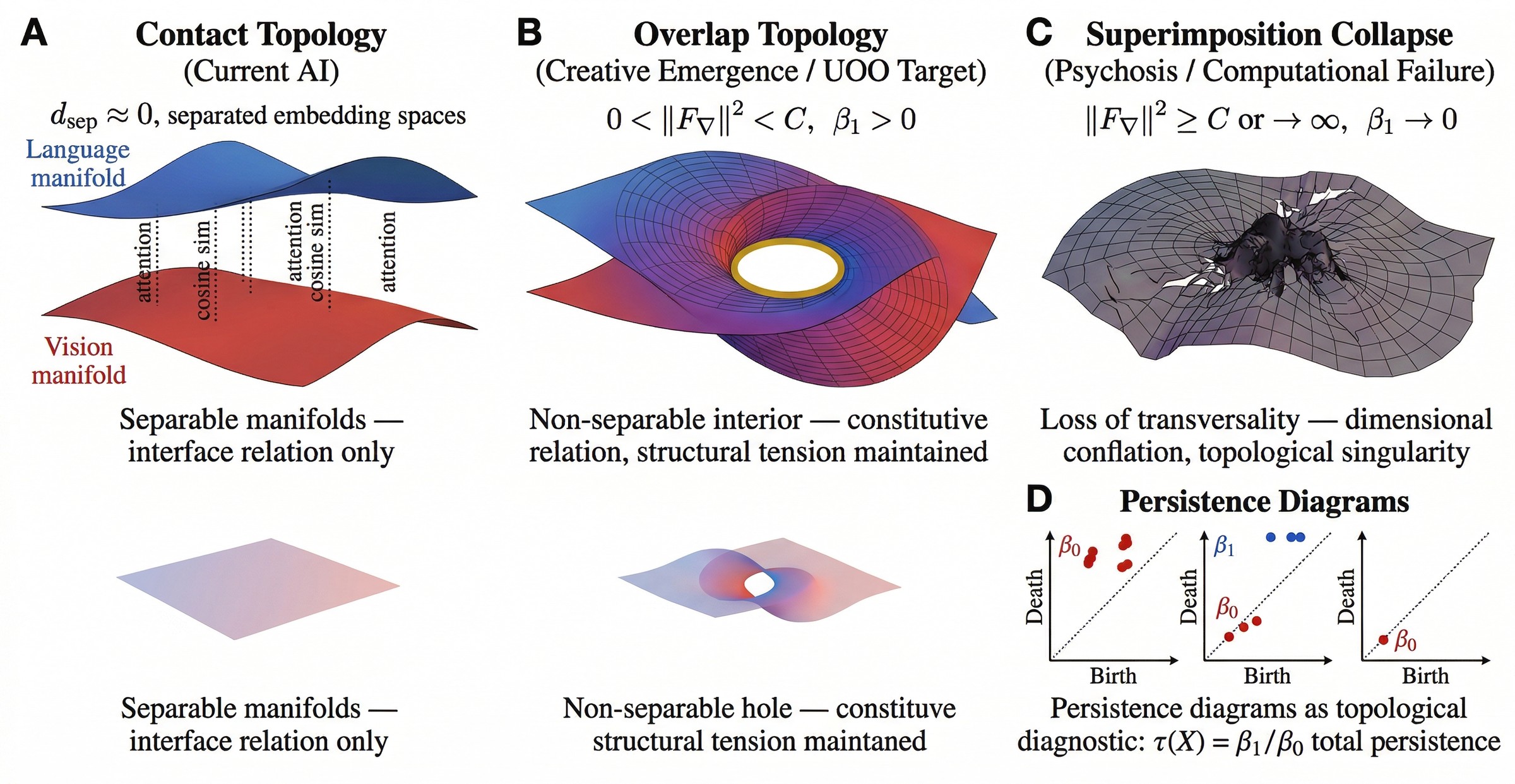}
\caption{\textbf{The Three Topological Regimes of Multimodal Integration.} Panel A: Contact topology (current AI)---separated manifolds with interface-only alignment. Panel B: Overlap topology (creative emergence / UOO target)---non-separable interior with persistent $\beta_1$ loops maintaining structural tension. Panel C: Superimposition collapse (psychosis / computational failure)---loss of transversality and topological singularity. Panel D: Corresponding persistence diagrams as quantitative diagnostic: $\tau(X) = \beta_1 / \beta_0$ total persistence.}
\label{fig:topological-regimes}
\end{figure}

Figure~\ref{fig:topological-regimes} visualizes this three-regime topology: Panel A
depicts the separated manifolds of contact topology, Panel B the
non-trivially entangled interior of overlap topology (with persistent
\(\beta_1\) loops), Panel C the singular collapse of superimposition,
and Panel D the corresponding persistence diagrams that serve as the
quantitative diagnostic.

\begin{center}\rule{0.5\linewidth}{0.5pt}\end{center}

\subsection{2. Related Work}\label{related-work}

\subsubsection{2.1 Multimodal Fusion
Architectures}\label{multimodal-fusion-architectures}

The evolution of multimodal AI can be traced through three generations.
Early approaches relied on feature-level concatenation or late fusion
(Ngiam et al., 2011), treating modalities as independent feature streams
merged only at the decision layer. The second generation, exemplified by
CLIP (Radford et al., 2021) and ALIGN (Jia et al., 2021), introduced
contrastive learning to align separate modal embedding spaces, achieving
remarkable zero-shot transfer but encoding separability as a structural
prior. The third and current generation---GPT-4V (OpenAI, 2023), Gemini
(Gemini Team, 2024), and Flamingo (Alayrac et al., 2022)---employs
cross-attention and interleaved architectures that achieve substantially
deeper modal integration. Yet as our analysis in §1.2 demonstrates, even
these systems preserve modality traceability as a residual separability
prior.

A parallel line of work in diffusion-based multimodal generation (Ramesh
et al., 2022; Saharia et al., 2022) has demonstrated that highly
entangled latent representations can emerge during the denoising
process, but training objectives systematically drive the system away
from these states. The present paper's contribution is to identify the
\emph{shared geometric prior} underlying all three generations and to
propose a mathematically grounded alternative.

\subsubsection{2.2 Topological Data Analysis in
Neuroscience}\label{topological-data-analysis-in-neuroscience}

The application of persistent homology to neural data has emerged as a
powerful tool for capturing structural features invisible to standard
statistical methods. Petri et al.~(2014) constructed homological
scaffolds of brain functional networks under psilocybin, revealing
persistent topological features that correlate with altered states of
consciousness. Saggar et al.~(2018) applied TDA to fMRI time series,
demonstrating that topological descriptors capture dynamical
organization of brain states during creative tasks more effectively than
linear dimensionality reduction. Stolz, Harrington, \& Porter (2017)
developed persistent homology methods for time-dependent functional
networks, providing tools for tracking topological evolution.

Our framework extends this line of research in two directions: (1) we
propose specific topological signatures (\(\beta_0\), \(\beta_1\),
\(\beta_2\) persistence profiles) that should \emph{discriminate}
between overlap and non-overlap processing states, generating
falsifiable predictions rather than post-hoc descriptions; (2) we
integrate persistent homology into the \emph{training objective} of
neural networks (§6.2), not merely as an analytical tool applied after
the fact.

\subsubsection{2.3 Geometric Deep
Learning}\label{geometric-deep-learning}

Bronstein et al.~(2021) unified convolutional, recurrent, graph, and
equivariant neural network architectures under a single geometric
framework organized by symmetry groups. Cohen et al.~(2019) demonstrated
gauge equivariant convolutional networks on manifolds, and Weiler \&
Cesa (2019) generalized steerable CNNs to arbitrary E(2)-equivariant
architectures. These advances establish the principle that architectural
design should begin with the symmetry structure of the processing
domain.

The present work extends GDL in a specific direction: from intra-modal
symmetry (e.g., rotational equivariance on images) to \emph{cross-modal}
symmetry on the modality configuration space. The Universal Overlap Zone
Operator (UOO) is a \(G\)-equivariant kernel on the fiber bundle section
space over the base space \(B\) of modality configurations---a domain
whose symmetry group \(G\) is itself an open problem.

\subsubsection{2.4 Graph Neural Networks and
Over-Squashing}\label{graph-neural-networks-and-over-squashing}

Gilmer et al.~(2017) formalized message passing neural networks; Alon \&
Yahav (2021) identified the over-squashing bottleneck in deep
GNNs---exponential information compression through fixed-dimension
message passing. Topping et al.~(2022) connected over-squashing to
discrete Ricci curvature, proposing curvature-based graph rewiring as a
remedy. Di Giovanni et al.~(2023) further analyzed the spectral
properties governing over-squashing.

Within our framework, over-squashing is the graph-level mechanism that
\emph{enforces} contact topology: it constrains cross-modal information
flow to pass through bottlenecks, preventing the formation of
non-separable representations. Anti-over-squashing techniques are thus
natural computational complements to fiber bundle connections (§5.6).

\subsubsection{2.5 Cognitive Neuroscience of
Creativity}\label{cognitive-neuroscience-of-creativity}

The neuroscience of creativity has undergone a paradigm shift from the
``right brain'' myth to network-level accounts. Beaty et al.~(2018)
established robust DMN-ECN co-activation as a neural signature of
creative ability. The subsequent tripartite framework incorporating the
Salience Network (Menon, 2011; Beaty, Seli, \& Schacter, 2019) provided
a mechanistic account of dynamic network coordination. Benedek \& Fink
(2019) proposed a neurocognitive framework emphasizing the role of
executive processes in creative idea generation.

Our contribution to this literature is threefold: (1) we provide a
\emph{topological} formalization of what ``co-activation'' means at the
representational level (inseparability, not mere co-occurrence); (2) we
reinterpret the SN as a coupling regulator operating in a
two-dimensional parameter space (coupling intensity × regulatory
capacity), compatible with Carson's (2011) shared vulnerability model;
(3) we generate differential predictions distinguishing the Overlap Zone
hypothesis from existing dual-process theories.

\subsubsection{2.6 Computational Psychiatry and Network
Models}\label{computational-psychiatry-and-network-models}

Adams et al.~(2013) proposed computational anatomy of psychosis based on
predictive coding frameworks. Anticevic et al.~(2012) demonstrated
aberrant default network deactivation in psychotic states.
Whitfield-Gabrieli \& Ford (2012) reviewed DMN hyperconnectivity in
schizophrenia. Palaniyappan \& Liddle (2012) argued for a cardinal role
of the salience network in psychosis.

The pathological mirror construct in our framework (§4) draws on and
extends this literature by proposing that the creativity-psychopathology
spectrum can be understood as occupying different regions of the same
parameter space, with SN gating integrity as the primary discriminating
variable. This reinterpretation generates specific testable predictions
(§4.3) not derivable from existing computational psychiatry models.

\subsubsection{2.7 Distinguishing the Present Framework from Related
Approaches}\label{distinguishing-the-present-framework-from-related-approaches}

Three lines of prior work share surface similarities with the present
framework but differ in fundamental ways that must be made explicit.

\textbf{Topological layers for classification (PLLay, PersLay).} Hofer
et al.~(2019) and Carrière et al.\ (2020) introduced
differentiable topological layers that compute persistence diagrams as
intermediate representations within neural networks. These methods
deploy TDA as a \emph{feature extraction} mechanism---topology is used
to read out structural properties of learned representations after the
fact. The present framework, by contrast, uses persistent homology
within the \emph{training objective itself}
(\(\mathcal{L}_{\text{topo}}\), §6.2): topological regularization
actively shapes the representational geometry during learning, rather
than passively describing it post-hoc. The distinction is between
topology-as-readout and topology-as-constraint.

\textbf{Topological Autoencoders (Moor et al., 2020).} Topological
Autoencoders preserve the topological structure of input data in the
latent space by penalizing discrepancies between input-space and
latent-space persistence diagrams. This is a single-domain
topology-preservation objective. Our framework's
\(\mathcal{L}_{\text{topo}}\) has a fundamentally different goal: it
does not preserve input topology but \emph{creates} specific topological
features (persistent \(\beta_1\) loops) in the cross-modal
representation space that do not exist in either input modality
separately. The target is cross-modal non-separability, not input-output
topological fidelity.

\textbf{Bayesian Program Learning (Lake et al., 2015).} BPL achieves
compositional concept learning through probabilistic programs over parts
and relations, enabling remarkable few-shot learning. However, BPL
operates entirely within a propositional-compositional representational
framework---concepts are decomposed into discrete, recombinant parts.
The Overlap Zone hypothesis claims that a class of cognitive operations
requires representations that are \emph{constitutively
non-decomposable}: the entangled representation cannot be reconstructed
from any combination of its component parts. BPL addresses combinatorial
generalization within separable representations; UOO addresses a
fundamentally different representational regime.

\begin{center}\rule{0.5\linewidth}{0.5pt}\end{center}

\subsection{\texorpdfstring{3. Philosophical Foundations: \emph{Xiàng}
as the Creative-Transformative
Zone}{3. Philosophical Foundations: Xiàng as the Creative-Transformative Zone}}\label{philosophical-foundations-xiuxe0ng-as-the-creative-transformative-zone}

The philosophical framework presented here is not a motivational
preamble to the mathematical and computational pillars---it is their
generative source. The fiber bundle formalism of §5 and the Neural ODE
architecture of §6 are the mathematical and computational
instantiations, respectively, of a philosophical structure discovered at
the intersection of Wittgenstein's analysis of language and the Chinese
craft epistemology tradition. This section constructs that structure
step by step.

\subsubsection{3.1 Wittgenstein's Problem: The Impassable Boundary
Between Saying and
Showing}\label{wittgensteins-problem-the-impassable-boundary-between-saying-and-showing}

In the \emph{Tractatus Logico-Philosophicus}, Wittgenstein drew a
boundary that has defined a century of philosophy: between what can be
\emph{said} (expressed in propositional form) and what can only be
\emph{shown} (manifested through the logical structure of propositions
themselves, yet inexhaustible by any finite set of propositions). This
is not a distinction between the intelligible and the mystically
unknowable---it is a precise claim about the \emph{structural boundary
of propositional representation itself}.

At the \emph{Tractatus}'s conclusion, Wittgenstein confronted the
consequences of his own analysis: ``What we cannot speak about we must
pass over in silence'' (7). This is typically read as a philosophical
humility gesture. But read structurally, it is a \emph{decision}: having
identified a domain that language can point toward but never fully
enter---the domain of what \emph{shows itself}---Wittgenstein chose
silence over operation. He saw two circles (saying and showing), drew
the line between them, and refused to enter---or could not see---the
zone of their overlap.

Yet Wittgenstein himself, at the \emph{Tractatus}'s very end, departed
from this purely logical stance: ``There are, indeed, things that cannot
be put into words. They \emph{show themselves} (\emph{Es zeigt sich}).
They are what is mystical'' (6.522). Here ``showing'' is no longer the
self-display of logical form---it is the self-arrival of \emph{das
Mystische} (the mystical). Wittgenstein touched something his own
philosophical framework could not accommodate---he sensed another
dimension of ``showing,'' one belonging not to logic but to existence
itself---but he chose silence.

The question that structures the present framework is: \textbf{Is
silence the only response to what cannot be fully said?} Or does there
exist a third mode---neither purely propositional nor purely
presentational---in which saying and showing interpenetrate to generate
something that neither could produce alone?

\subsubsection{\texorpdfstring{3.2 The Chinese Response: \emph{Xiàng} as
the Third
State}{3.2 The Chinese Response: Xiàng as the Third State}}\label{the-chinese-response-xiuxe0ng-as-the-third-state}

The Chinese epistemological tradition provides a direct answer to the
question Wittgenstein left open. The \emph{Yijing} tradition's
foundational methodological statement---``The sage establishes
\emph{xiàng} to express meaning fully'' (shengren li xiang yi jin yi, \emph{Xici
zhuan})---is not a parallel formulation of Wittgenstein's distinction.
It is a \textbf{response to his silence, issued from a tradition he
could not have known}.

At the point where language cannot exhaust meaning (\emph{shū bù jìn
yán, yán bù jìn yì}---``writing cannot exhaust speech, speech cannot
exhaust meaning''), the Chinese tradition does not fall silent. It
\emph{operates}: the sage ``establishes \emph{xiàng}''---constructs
images. \emph{Xiàng} (xiang) is not a static picture but an
\textbf{operative schema}: a dynamic structure that simultaneously
organizes perceptual, conceptual, and procedural knowledge without being
reducible to any one of these dimensions.

The critical insight, crystallized in the analysis presented in the
companion document (Tan, 2026b), is that \emph{xiàng} is \textbf{neither
purely saying nor purely showing, but the third state that emerges when
saying and showing cross each other's boundaries}:

\begin{itemize}
\tightlist
\item
  \emph{Xiàng} is not a propositional statement---it cannot be fully
  paraphrased.
\item
  \emph{Xiàng} is not a mere percept---it is not a mental image waiting
  to be decoded.
\item
  \emph{Xiàng} is what happens when the propositional and the
  presentational \emph{interpenetrate}: a meaning-emergence event in
  which both modes are simultaneously operative and structurally
  inseparable.
\end{itemize}

An architectural blueprint exemplifies this structure: it simultaneously
\emph{says} spatial relationships (dimensions, structural loads,
regulatory compliance---information that can be read and transmitted
linguistically) and \emph{shows} a phenomenological quality (spatial
experience, light, proportion---content that must be \emph{seen} to be
understood). This dual character is not a deficiency; it is the
blueprint's creative power. The plan drawing is not a record of a
building that already exists---it is the event in which a new thing
comes into being \emph{precisely at the intersection of saying and
showing}.

\subsubsection{\texorpdfstring{3.3 The Cruciform Structure:
\emph{Dào}/\emph{Qì} ×
Saying/Showing}{3.3 The Cruciform Structure: Dào/Qì × Saying/Showing}}\label{the-cruciform-structure-duxe0oquxec-sayingshowing}

The saying/showing distinction provides the horizontal axis of the
framework. The vertical axis comes from a different philosophical
tradition: the Chinese cosmological distinction between \emph{dào} (dao,
the formless, the metaphysical) and \emph{qì} (qi, the formed, the
physical). Superimposing these two axes yields a \textbf{cruciform
structure}:

\begin{verbatim}
                    dao (dào — the formless Way)
                         ↑
                         |
    yanshuo (saying) ←——  tu-xiang (xiàng)  ——→ xianshi (showing)
                         |
                         ↓
                    qi (qì — formed artifact)
\end{verbatim}

\emph{Xiàng} occupies the intersection of the cross. It simultaneously
executes two operations of \emph{huàcái} (huacai,
transformation-and-cutting):

\begin{itemize}
\item
  \textbf{Vertical \emph{huàcái}}: Mediating between \emph{dào} and
  \emph{qì}---how the formless Way descends into formed artifact. This
  was the core concern of the author's doctoral thesis on architectural
  image philosophy (Tan, 2008), examining how craft traditions from the
  \emph{Mingtang} (Mingtang, Bright Hall) system to the \emph{Yingzao Fashi}
  (Yingzao Fashi, Building Standards) encode the passage from cosmic
  principle to material construction.
\item
  \textbf{Horizontal \emph{huàcái}}: Mediating between saying and
  showing---establishing passage between what language can capture and
  what can only be directly presented. This is the Wittgensteinian
  dimension, and the one that connects the philosophical framework
  directly to the topological diagnosis of multimodal AI.
\end{itemize}

The cruciform structure provides a precise coordinate system for
locating different types of image-operation within the Chinese craft
tradition:

\begin{itemize}
\tightlist
\item
  The \emph{Mingtang} system tends toward the \emph{dào}--showing
  quadrant: it is a direct cosmological presentation, more shown than
  said.
\item
  The \emph{Yingzao Fashi} tends toward the \emph{qì}--saying quadrant:
  it encodes craft knowledge as transmissible textual rules.
\item
  The \emph{Jiuzhang Suanshu} (Jiuzhang Suanshu, Nine Chapters on the
  Mathematical Art) sits at the saying--showing intersection: its
  \emph{chū-rù xiāng-bǔ} method simultaneously relies on textual
  reasoning and visual intuition.
\item
  The \emph{Yijing} occupies the exact center of the cross: its hexagram
  generation is simultaneously a saying (guaci/yaoci, hexagram/line
  texts), a showing (guaxiang, hexagram images), a cosmological embodiment
  (\emph{dào} as \emph{tàijí}), and a normative artifact (\emph{qì} as
  ritual instrument).
\end{itemize}

\subsubsection{\texorpdfstring{3.4 \emph{Chuànghuà} and \emph{Huàcái}:
The Dual-Layer Dynamics of the Creative-Transformative
Zone}{3.4 Chuànghuà and Huàcái: The Dual-Layer Dynamics of the Creative-Transformative Zone}}\label{chuuxe0nghuuxe0-and-huuxe0cuxe1i-the-dual-layer-dynamics-of-the-creative-transformative-zone}

The cruciform structure reveals a dual-layer dynamic at its intersection
point that resolves a long-standing ambiguity in image theory: are
images creative events or normative tools?

\textbf{The answer is both, in sequence:}

\textbf{\emph{Chuànghuà} (chuanghua, creative transformation)} is a
generative event---it does not make decisions; it grows. At the
intersection of saying and showing, \emph{xiàng} emerges spontaneously,
like a vortex at the confluence of two rivers. This is the moment Cook
Ding's knife finds the space between joints ``by spirit and not by
sight'' (\emph{Zhuangzi}): the image of cutting arises not from the
cook's intention but from the encounter between his embodied skill
(showing) and the ox's anatomical logic (saying). \emph{Chuànghuà} is
one-time, unrepeatable, irreducible.

\textbf{\emph{Huàcái} (huacai, institutional cutting)} is the
systematization of \emph{chuànghuà}---when the creative event is
captured, encoded, and made available for repeated use. Cook Ding's
knife-art, once it enters the \emph{Zhuangzi} as a parable, becomes
\emph{huàcái}---a transmissible pattern. When the \emph{Yingzao Fashi}
codifies centuries of craft intuition into dimensional standards and
modular ratios, it performs \emph{huàcái} on accumulated
\emph{chuànghuà}.

The relationship between the two is not opposition but \textbf{phase
transition}: the same \emph{xiàng}, at the moment of its first
emergence, is \emph{chuànghuà}; once encoded into the
\emph{zhì-qì-zhī-lǐ} value chain (``from raw material to finished
artifact''), it becomes \emph{huàcái}. Every craftsman who builds by the
code re-enacts, in institutionalized form, the original creative event.
\emph{Huàcái} contains \emph{chuànghuà} as its remembered origin;
\emph{chuànghuà} generates \emph{huàcái} as its transmissible form.

This dual-layer structure maps directly onto the computational
framework:

\begin{itemize}
\tightlist
\item
  \emph{Chuànghuà} corresponds to the Neural ODE trajectory
  \(z(0) \to z(T)\)---the continuous, dynamic process in which
  non-trivial \(\beta_1\) topology emerges in representation space.
\item
  \emph{Huàcái} corresponds to the frozen parameters \(\theta\) after
  training---the institutionalized operator that can be repeatedly
  applied to new inputs.
\item
  Superimposition collapse corresponds to the failure of
  \emph{chuànghuà}---the creative event degenerates into noise, just as
  misfolded proteins lose functional structure.
\end{itemize}

\subsubsection{3.5 From Philosophical Structure to Architectural
Prediction}\label{from-philosophical-structure-to-architectural-prediction}

The cruciform framework generates a precise architectural prediction
more powerful than the one stated in v5.0. The ceiling of current
multimodal AI is not merely ``topological rather than parametric'' in a
generic sense---it can now be diagnosed with structural specificity:

\textbf{Diagnosis}: Current architectures suppress the Overlap Zone
because their training objectives enforce a regime in which
\emph{saying} and \emph{showing} remain on opposite sides of a
boundary---contact topology is the computational implementation of
Wittgenstein's silence. Contrastive alignment forces each modality to
\emph{stay in its circle}; cross-attention allows information to cross
the boundary but preserves \emph{which circle it came from} (modality
traceability); denoising objectives penalize \emph{dwelling} at the
intersection.

\textbf{Prediction}: Any system whose training objectives do not
explicitly maintain and reward representations at the saying--showing
intersection---representations with non-trivial \(\beta_1\) topology in
the persistence diagram of the cross-modal representation space---will
systematically fail on tasks requiring \emph{chuànghuà}-type processing.
This includes tasks requiring deep cross-modal metaphor, structural
analogy across disparate domains, and the kind of implicit creativity
that master craftsmen demonstrate when operating in the
\emph{xiàng}-generating mode.

\textbf{Falsification}: If highly creative individuals and those with
high schizotypy exhibit topologically indistinguishable \(\tau(X)\)
signatures---i.e., if the \emph{chuànghuà}/\emph{collapse} distinction
does not correspond to a detectable topological difference---then the
philosophical framework's core claim collapses, and the program exits.

\begin{quote}
The full conceptual genealogy of \emph{xiàng}---including the original
Venn diagram analysis, the cruciform structure's derivation, and the
\emph{chuànghuà}/\emph{huàcái} double-layer dynamics with historical
case studies from the \emph{Mingtang}, \emph{Yingzao Fashi},
\emph{Jiuzhang Suanshu}, and \emph{Yijing}---is developed in the
companion monograph (Tan, 2026b). The present section condenses this
argument to its structural essentials.
\end{quote}

\begin{center}\rule{0.5\linewidth}{0.5pt}\end{center}

\subsection{4. Cognitive Science Pillar: The Neural Basis of the Overlap
Zone}\label{cognitive-science-pillar-the-neural-basis-of-the-overlap-zone}

\subsubsection{4.1 The Co-Activation Puzzle in Creativity
Neuroscience}\label{the-co-activation-puzzle-in-creativity-neuroscience}

Beaty et al.~(2018, \emph{PNAS}) demonstrated that highly creative
individuals exhibit abnormally high functional connectivity between the
Default Mode Network (DMN) and the Executive Control Network (ECN)---two
systems that typically suppress each other---during creative tasks. This
finding has been widely cited and rigorously replicated, yet \emph{why
this co-activation produces creative outcomes rather than mere cognitive
conflict} remains theoretically unexplained.

The Overlap Zone framework offers a concrete candidate mechanism:
DMN-ECN co-activation during creative cognition is not mere ``network
crosstalk'' but the neural correlate of entering a specific
representational state---the Overlap Zone---in which non-propositional
content (DMN-associated imagery, synesthetic association) and
propositional content (ECN-associated executive control, logical
construction) are simultaneously present and structurally inseparable.

\subsubsection{4.2 The Tripartite Network Framework: From Dual Network
to
DMN/ECN/SN}\label{the-tripartite-network-framework-from-dual-network-to-dmnecnsn}

Recent creativity neuroscience has advanced from a dual-network model
(DMN vs.~TPN) to a tripartite framework incorporating the
\textbf{Salience Network} (SN), anchored in the dorsal anterior
cingulate cortex (dACC) and anterior insula (aINS) (Menon, 2011; Beaty,
Seli, \& Schacter, 2019).

The SN plays a critical gating role: it detects behaviorally relevant
conflicts between internally generated signals (DMN) and goal-directed
signals (ECN), dynamically modulating the coupling strength between the
two networks. Within the Overlap Zone framework, the SN functions not as
a simple ``switch'' (toggling between DMN and ECN modes) but as a
\textbf{coupling regulator}---determining the degree to which
propositional processing (ECN) and non-propositional processing (DMN)
are permitted to enter a state of inseparable co-constitution.

This interpretation generates a key theoretical distinction---the
Overlap Zone framework must adjudicate between two competing models:

\begin{itemize}
\tightlist
\item
  \textbf{Optimal coupling model} (graded): Creative performance is a
  continuous function of DMN-ECN coupling strength, peaking at an
  optimal value (consistent with the inverted-U finding of Acar \& Sen,
  2013). SN-mediated coupling is not an all-or-nothing switch but
  dynamic regulation maintaining an \textbf{optimal coupling window}.
\item
  \textbf{Inseparable processing state model} (threshold): A qualitative
  transition point exists above which representations shift from
  approximately separable to genuinely inseparable---the Overlap Zone
  ``activates.''
\end{itemize}

The two are not mutually exclusive: the system may exhibit graded
differences after crossing the inseparability threshold. The
dual-threshold framework in §5 (Conjecture 3) mathematically encodes
this compatibility.

\subsubsection{4.3 Inseparability Signature and Differential
Predictions}\label{inseparability-signature-and-differential-predictions}

The core experimental prediction of the Overlap Zone framework is the
\textbf{inseparability signature}: in creative cross-modal tasks,
selective perturbation of one dimension should produce semantically
related synchronous changes in the other dimension, and vice versa.

This prediction is more specific than general dual-task interference.
Dual-task interference predicts that perturbing one dimension will
\emph{impair} the other (resource competition). The inseparability
signature predicts a stronger effect: perturbation will produce
\emph{semantically related covariation}---a content change in one
dimension that is structurally related to the perturbation content in
the other. This distinction enables falsificational discrimination
against existing dual-process theories.

\textbf{Differential prediction}: If the Overlap Zone theory holds, high
DMN-ECN functional connectivity should not be uniformly elevated across
all ``creative tasks''---it should show a disproportionate surge in
tasks that strictly require inseparability (e.g., cross-modal metaphor
construction) while remaining significantly lower in creative tasks
solvable by sequential modality-specific processing (e.g., purely verbal
creativity tasks lacking a simultaneous visual-spatial dimension). This
differential prediction goes beyond the coarse-grained descriptions of
resource competition or sequential binding in existing dual-process
accounts.

\textbf{Three-way experimental discrimination.} The experimental design discriminates three mutually exclusive outcomes, each with distinct behavioral, neural, and computational signatures:

\begin{enumerate}
\item \textbf{Outcome A: Correlated degradation (Overlap Zone prediction).}
\begin{itemize}
\item \emph{Behavioral}: Suppressing propositional processing (e.g., articulatory suppression) degrades \emph{both} verbal fluency \emph{and} non-propositional imagery quality, with Pearson $r > 0.5$ between degradation magnitudes across dimensions.
\item \emph{Neural}: Simultaneous decrease in both DMN and ECN activation during suppression, with preserved DMN-ECN functional connectivity ($\Delta\mathrm{FC} < 10\%$). The Salience Network shows compensatory upregulation.
\item \emph{Computational}: Perturbation along one modal axis produces correlated displacement along the orthogonal axis---the perturbation Jacobian $\partial z_{\text{showing}} / \partial z_{\text{saying}}$ is full-rank, indicating constitutive entanglement.
\item \emph{Topological}: $\tau(X)$ decreases uniformly under suppression; $\beta_1$ persistence declines in proportion to suppression strength.
\end{itemize}

\item \textbf{Outcome B: Trade-off (Separable, shared-resource prediction).}
\begin{itemize}
\item \emph{Behavioral}: Suppressing one dimension \emph{improves} the other---articulatory suppression releases resources that enhance imagery/spatial processing. Negative cross-dimensional correlation ($r < -0.3$).
\item \emph{Neural}: Decreased activation in the suppressed network with \emph{increased} activation in the complementary network. Anti-correlated network dynamics.
\item \emph{Computational}: Rank-1 perturbation---displacement along one axis with zero or opposite displacement along the orthogonal axis. The perturbation Jacobian is rank-deficient.
\item \emph{Topological}: $\beta_0$ persistence (disconnected components) increases under suppression as the representation separates into modality-specific clusters.
\end{itemize}

\item \textbf{Outcome C: No cross-dimensional effect (Separable, independent-channels prediction).}
\begin{itemize}
\item \emph{Behavioral}: Suppressing one dimension affects only that dimension; the other remains at baseline. Near-zero cross-dimensional correlation ($|r| < 0.1$).
\item \emph{Neural}: Suppression affects only the targeted network with no measurable change in the complementary network or their functional connectivity.
\item \emph{Computational}: Block-diagonal perturbation structure---each modal subspace is independently modifiable.
\item \emph{Topological}: No change in $\tau(X)$ or $\beta_1$ persistence; the cross-modal topological structure was never constitutive.
\end{itemize}
\end{enumerate}

\noindent Only Outcome~A is consistent with the Overlap Zone hypothesis. Outcomes B and C are both consistent with separable processing but distinguishable from each other by the sign of the cross-dimensional correlation. This three-way discrimination is the framework's strongest falsifiability feature: the theory specifies not merely what should happen (Outcome~A) but precisely what \emph{should not} happen (Outcomes B and C), with quantitative thresholds for each.

\subsubsection{4.4 Pathological Mirror: Overlap Isomorphism
vs.~Superimposition
Collapse}\label{pathological-mirror-overlap-isomorphism-vs.-superimposition-collapse}

\paragraph{4.4.1 The Topological
Opposition}\label{the-topological-opposition}

In establishing the ontological status of the Overlap Zone, a rigorous
conceptual clarification is required: ``overlap'' (intersection) and
``superimposition'' (collapse) represent diametrically opposed dynamical
outcomes within the topological and cognitive-pathological framework of
this research. This distinction is not merely terminological---it
constitutes the theory's \textbf{falsification condition}: the theory
must simultaneously predict its own failure mode.

\textbf{Topological dimension: Maintenance vs.~loss of transversality.}
In algebraic geometry, \textbf{transversal intersection} preserves the
local structural integrity of each participating object---the tangent
spaces at intersection points jointly span the ambient space.
Non-transversal intersection degenerates local structure, with
dimensional information lost. The Overlap Zone corresponds to
\textbf{transversal coupling} of propositional and non-propositional
dimensions: the two dimensions maintain their respective structural
integrity within a shared interior region, giving rise to creative
transformation (\emph{chuànghuà}). Superimposition collapse corresponds
to \textbf{loss of transversality}: dimensional distinction dissolves
and generative tension is destroyed.

\paragraph{4.4.2 The Cognitive-Pathological
Dimension}\label{the-cognitive-pathological-dimension}

Extensive meta-analytic evidence (Acar \& Sen, 2013; Kyaga et al., 2011;
Taylor, 2017) establishes a systematic but nonlinear association between
creativity and psychopathology-spectrum traits. Carson's (2011) shared
vulnerability model argues that creativity and psychopathology may share
specific cognitive features (cognitive disinhibition, associative
loosening), but that protective factors such as high working memory
capacity determine whether these features lead to creative output or
clinical symptoms.

Within the Overlap Zone framework, this picture can be reinterpreted:
the Overlap Zone and Superimposition Collapse are not two endpoints of
the same dimension, but two distinct regions in an at least
\textbf{two-dimensional parameter space}---one dimension being
\textbf{coupling intensity} (DMN-ECN coupling strength) and the other
being \textbf{regulatory capacity} (SN gating integrity, working memory
capacity, etc.). Creative emergence occupies the quadrant of high
coupling and high regulation; cognitive disorganization occupies high
coupling with low regulation. This two-dimensional model is compatible
with the inverted-U relationship between creativity and unusual
cognitive experiences reported in meta-analyses (Acar \& Sen, 2013):
when regulatory capacity is insufficient, excessive coupling intensity
no longer enhances creativity but leads to disorganization.

\paragraph{4.4.3 SN Gating Failure and Superimposition
Collapse}\label{sn-gating-failure-and-superimposition-collapse}

The watershed between the Overlap Zone and Superimposition Collapse lies
primarily not in the intensity of DMN-ECN interaction itself, but in the
\textbf{integrity of SN gating function}. The Overlap Zone is the
dynamical regime in which the SN successfully coordinates high-intensity
DMN-ECN coupling; Superimposition Collapse is the regime of deregulated
DMN-ECN coupling following SN gating deterioration. This
reinterpretation aligns with growing evidence in the schizophrenia
literature that structural and functional abnormalities in the dACC and
aINS may be upstream drivers of network dynamics disruption (Menon,
2011; Palaniyappan \& Liddle, 2012; Whitfield-Gabrieli \& Ford, 2012;
Anticevic et al., 2012).

\textbf{Testable predictions:}

\begin{enumerate}
\def\labelenumi{\arabic{enumi}.}
\tightlist
\item
  In the overlap regime, SN node activity should positively correlate
  with the \textbf{rate of change} of DMN-ECN coupling strength---the SN
  actively modulating instantaneous coupling dynamics.
\item
  In the superimposition collapse regime, this correlation should
  \textbf{vanish or reverse}---the SN no longer effectively modulating
  coupling fluctuations.
\item
  Highly creative individuals and those with high schizotypy scores may
  show no significant difference in DMN-ECN coupling \textbf{intensity},
  but should differ significantly in SN gating \textbf{integrity} (i.e.,
  the correlation coefficient from prediction 1).
\end{enumerate}

\begin{quote}
\textbf{Causal direction caveat}: The above predictions are based on
correlational designs. Existing evidence is equally compatible with a
reverse causal chain (cognitive processes drive network reorganization →
topological signature changes). Causal tests---e.g., temporarily
suppressing SN function via transcranial magnetic stimulation (TMS) and
observing whether creative thought quality correspondingly
degrades---are planned as long-term goals in Phase 4, not as near-term
experimental components.
\end{quote}

\paragraph{4.4.4 Computational Extension: Scope and
Limits}\label{computational-extension-scope-and-limits}

An analogy previously proposed must be explicitly rejected as
technically incorrect: LLM hallucination is \textbf{not} superimposition
collapse. LLMs do not possess the propositional/non-propositional
dual-dimension structure presupposed by the Overlap Zone
framework---every representation is a vector in the same embedding
space, with no structural boundary that could ``collapse.'' LLM
hallucination arises from distributional gaps and calibration
failure---problems at the probability distribution level, not
topological events.

However, a more precise computational prediction exists: a multimodal
system trained with a UOO architecture---designed to maintain
non-separable cross-modal representations---may, under certain stress
conditions, see its cross-modal representations collapse from the
overlap regime to the superimposition regime (§7.1 details three
specific stress conditions). This constitutes a novel prediction about a
system that does not yet exist.

\begin{center}\rule{0.5\linewidth}{0.5pt}\end{center}

\subsection{5. Mathematical Pillar: Fiber Bundles and
Connections}\label{mathematical-pillar-fiber-bundles-and-connections}

\subsubsection{5.1 The Overlap Zone Isomorphism
Conjecture}\label{the-overlap-zone-isomorphism-conjecture}

We conjecture that the Overlap Zone processing structure---the
inseparable co-constitution of propositional and non-propositional
content---is \textbf{invariant across modality pairs}. The
visual-language, auditory-motor, and spatial-conceptual Overlap Zones,
despite relying on different neural substrates and processing different
content types, share the same structural characteristics.

If this conjecture holds, it explains how creativity transfers across
domains: the structural invariant that masters carry between media
(\emph{xiàng}) is not a domain-specific skill but a universal processing
structure that manifests in the same form regardless of the specific
modalities involved.

\subsubsection{5.2 The Fiber Bundle
Framework}\label{the-fiber-bundle-framework}

We propose connections on fiber bundles as the mathematical language for
formalizing cross-domain structural invariance:

\begin{itemize}
\tightlist
\item
  \textbf{Base space} \(B\): The space of modality configurations---all
  possible modality pairs.
\item
  \textbf{Fiber} \(E_b\): For each modality pair \(b \in B\), the
  manifold of all possible Overlap Zone representations under that
  modality pair.
\item
  \textbf{Connection} \(\nabla\): Defines parallel transport between
  fibers---how to move Overlap Zone representations across modality
  pairs while preserving structural characteristics.
\item
  \textbf{Curvature} \(F_\nabla\): Measures the path-dependence of
  parallel transport. Zero curvature means cross-domain isomorphism
  holds exactly; nonzero curvature means the specific modal path
  influences the transported representation.
\end{itemize}

\emph{Xiàng} receives a precise identification in this framework: it is
a \emph{stable section} of this fiber bundle---an entity whose
structural characteristics remain invariant under parallel transport
across modality pairs.

\paragraph{\texorpdfstring{5.2.1 Candidate Geometric Constructions for
\(B\)}{5.2.1 Candidate Geometric Constructions for B}}\label{candidate-geometric-constructions-for-b}

\textbf{Candidate Path A: Grassmannian (near-term default).} Let
\(\mathbb{C}^n\) be a high-dimensional signal representation Hilbert
space; each modality defines a \(k\)-dimensional subspace. The space of
all modality configurations is
\(B = \text{Gr}(k, \mathbb{C}^n) \times \text{Gr}(k, \mathbb{C}^n)\), a
compact smooth manifold of dimension \(2k(n-k)\) with a natural
U(\(n\))-action. \textbf{Advantage}: Concretely computable, with
well-studied topological invariants (Chern classes).
\textbf{Limitation}: Requires pre-specifying \(k\) and \(n\).

\textbf{Candidate Path B: Information geometry (long-term alternative).}
Define each modality as a statistical manifold \(\mathcal{M}_i\)
equipped with the Fisher-Rao metric. The geometry of \(B\) is then
data-derived rather than theoretically imposed. \textbf{Advantage}:
Data-driven. \textbf{Limitation}: Fisher-Rao metric may be
computationally intractable in high dimensions; if \(B\) is effectively
a finite discrete set (e.g., only three modality pairs), standard
differential structure does not exist, requiring discrete connection
theory---a significant mathematical complication.

The two paths are non-exclusive. Near-term computational validation
(§6.3) uses Path A; Path B provides long-term insurance for discovering
structures that Path A misses.

\paragraph{\texorpdfstring{5.2.2 Structure Group \(G\): Recommended
Candidate and Open
Problem}{5.2.2 Structure Group G: Recommended Candidate and Open Problem}}\label{structure-group-g-recommended-candidate-and-open-problem}

\textbf{Recommended candidate}: \(G = \text{U}(n)\), the
\(n\)-dimensional unitary group. Rationale: (1) U(\(n\)) is the natural
acting group of Grassmannians, preserving the Fubini-Study metric; (2)
U(\(n\)) guarantees that non-separability entropy (NS) is invariant
under gauge transformations---a core physical requirement.

However, the final determination of \(G\) is the mathematical program's
\textbf{first-priority open problem}. \(G\) defines which fiber
transformations are permissible---it precisely delimits what
``cross-modal invariance'' means. \textbf{The determination of \(G\)
logically precedes the proof of all three conjectures.}

\textbf{Data-driven discovery path}: If \(G = \text{U}(n)\) proves too
large or too small, equivariance discovery methods can extract the
approximate structure of \(G\) from cross-modal task representation
data. This aligns with the latest direction in GDL research: learning
domain symmetry groups from data rather than prior specification.

\subsubsection{5.3 Three Conjectures (Nested
Hierarchy)}\label{three-conjectures-nested-hierarchy}

\begin{quote}
\textbf{Editor's note}: The following three propositions are advanced as
\emph{conjectures}. Their key constitutive objects---\(B\), \(G\),
\(E_b\)---have not yet achieved the precision required for
mathematicians to initiate formal proof work. They are proposed to
orient the mathematical program, generate falsifiable predictions, and
invite collaborators to participate in reformulation. Reformulation is
as valuable as solution.
\end{quote}

\paragraph{Conjecture 1: Existence --- Three-Level Nested
Formulation}\label{conjecture-1-existence-three-level-nested-formulation}

Let \(\Omega\) be an operator satisfying the three UOO conditions
(maintaining inseparability, bidirectional semantic coupling,
cross-domain structural invariance):

\textbf{Conjecture 1a (strong --- variational principle):} \(\Omega\)
derives via a variational principle from the curvature \(F_\nabla\) of
connections on a finite-dimensional principal bundle \(P(M,G)\)---i.e.,
\(\Omega\) is a solution to the critical point condition of an action
functional \(S[\nabla]\). \emph{Mathematical implication}: If true, UOO
is governed by Euler-Lagrange-type equations, drastically constraining
the search space and potentially admitting analytic solutions.

\textbf{Conjecture 1b (medium --- local functional):} \(\Omega\) is a
local functional of \(F_\nabla\) constrained by gauge equivariance.
\emph{Mathematical implication}: Corresponds to the local receptive
field property of GDL convolutions; directly mappable to a
gauge-equivariant computational prototype.

\textbf{Conjecture 1c (weak --- pure existence):} There exists a
finite-dimensional fiber bundle from whose connection and curvature
\(\Omega\) can be recovered, with no restriction on functional type.

\textbf{Reduced-dimension version}: \(B = S^2 \times S^2\),
\(G = \text{U}(1)\). For which Chern numbers do U(1)-connections
satisfying the UOO conditions exist? This connects directly to harmonic
maps (Jost, 2017) and gauge field theory, providing a tractable
mathematical entry point.

\paragraph{Conjecture 2: Isomorphism --- Two Directional
Propositions}\label{conjecture-2-isomorphism-two-directional-propositions}

\textbf{Conjecture 2-forward:} If modality pairs \(A\) and \(B\) have
diffeomorphic modality configuration manifolds, then there exists a
curvature-preserving gauge transformation \(T\) such that
\(T(\Omega_A) = \Omega_B\). \emph{Near-term operability}: Constructively
finding \(T\) is the path to zero-shot cross-modal transfer.

\textbf{Conjecture 2-reverse:} If such a \(T\) exists, then the
configuration manifolds of \(A\) and \(B\) are diffeomorphic.
\emph{Theoretical depth}: Analogous to Donaldson invariants extracting
differential-topological information from gauge theory; depends on
detailed properties of \(G\) and \(M\).

\begin{quote}
\textbf{Remark (Gauge symmetry and cross-modal transfer).}
The forward direction of Conjecture~2 deploys gauge symmetry in the standard sense of principal bundle theory. A gauge transformation is an automorphism of the total space $P$ that projects to the identity on the base $B$---a $G$-equivariant map $\Phi: P \to P$ covering $\mathrm{id}_B$. In the present setting, we extend this to automorphisms of $B$ itself: diffeomorphisms $\phi: B \to B$ that permute modality configurations while preserving the bundle structure. Such maps lift to bundle isomorphisms $\tilde{\phi}: P \to \phi^* P$, and the conjecture asserts that if $\phi$ is a diffeomorphism, then the pulled-back connection $\phi^* \nabla$ is gauge-equivalent to $\nabla$---i.e., the curvature tensor $F_\nabla$, which encodes the overlap zone's structural invariants, is preserved up to conjugation by $G$.

\emph{Computational implication.} If the forward direction holds, a UOO operator trained on one modality pair $(m_1, m_2)$ can transfer to a diffeomorphic pair $(m_1', m_2')$ without retraining---the gauge transformation provides the explicit transfer map. This is the fiber-bundle formalization of zero-shot cross-modal transfer: the creative-cognitive structure learned in one domain is not merely analogous but \emph{gauge-equivalent} to that in another.

\emph{Donaldson-type depth of the reverse.} The reverse direction is substantially deeper, analogous to Donaldson's theorem (1983) that smooth structures on 4-manifolds can be distinguished by Yang-Mills instantons: the moduli space of anti-self-dual connections on $P$ carries differential-topological information about the base that is invisible to classical invariants. In our setting, the reverse conjecture would mean that curvature data alone---the pattern of cross-modal interactions encoded in $F_\nabla$---suffices to recover the diffeomorphism type of the modality configuration space. This would establish that the overlap zone's topological invariants are not merely \emph{compatible with} but \emph{fully determine} the structure of modal relationships. Computing the relevant Donaldson invariants for the modality bundle is a program we defer to future work (\S10.2), with the reduced-dimension model ($B = S^2 \times S^2$, $G = \mathrm{U}(1)$) of Conjecture~1 as the analytically tractable entry point.
\end{quote}

\paragraph{Conjecture 3: Non-Separability Characterization ---
Dual-Threshold
Model}\label{conjecture-3-non-separability-characterization-dual-threshold-model}

Define the distance from representation \(z\) to the nearest separable
tensor:

\[d_{\text{sep}}(z) = \inf_{u,v} \|z - u \otimes v\|\]

Two critical thresholds \(\delta^* < \delta^{**}\) demarcate three
functional regimes:

\begin{itemize}
\tightlist
\item
  \(d_{\text{sep}} \approx 0\): Separable representation (contact
  topology, current AI default)
\item
  \(\delta^* < d_{\text{sep}} < \delta^{**}\): Overlap Zone (structural
  tension maintained, creative emergence)
\item
  \(d_{\text{sep}} > \delta^{**}\): Superimposition collapse
  (deregulated entanglement, cognitive disorganization)
\end{itemize}

\textbf{Operationalization}: Computing \(d_{\text{sep}}\) in
high-dimensional tensor spaces is intractable (NP-hard). We introduce
the Shannon entropy of Schmidt decomposition coefficients as a
computationally feasible proxy:

\[\text{NS}(z) = -\sum_i p_i \log p_i\]

where \(p_i\) are the squared Schmidt decomposition coefficients of
representation \(z\). NS \(= 0\) iff \(z\) is separable. The dual
thresholds map in NS-space: \(\text{NS} < \kappa^*\) (contact regime),
\(\kappa^* \leq \text{NS} < \kappa^{**}\) (overlap regime),
\(\text{NS} \geq \kappa^{**}\) (collapse regime). Here \(\kappa^{**}\)
corresponds to the downward inflection point of the inverted-U
creativity-schizotypy relationship, and is self-consistent with the
critical curvature threshold \(C\) in the Yang-Mills three-regime model
(§5.7).

\textbf{Operationalized research question}: Define
\(\varepsilon(\kappa) = \inf_{\text{separable } S} \sup_{(x,y)} \|\Omega(x,y) - S(x,y)\|\).
(a) Is \(\varepsilon(\kappa)\) monotonically increasing in \(\kappa\)?
(b) Does a critical \(\kappa^* > 0\) exist such that \(\varepsilon = 0\)
for \(\kappa < \kappa^*\) and \(\varepsilon > 0\) for
\(\kappa > \kappa^*\)? (c) If \(\kappa^*\) exists, does it coincide with
empirically observed inseparability signature thresholds? This question
connects directly to entanglement distillation thresholds (Horodecki et
al., 2009), nuclear norms, and tensor approximation theory---and is
recommended as the most self-contained mathematical entry point for
collaborators.

\textbf{Empirical \(\kappa^*\) detection protocol.} The phase-transition characterization of the dual thresholds is not merely assumed---it generates a concrete empirical signature. During the PoC's \(\alpha\)-sweep (§6.3), the empirical NS distribution over trained representations should exhibit one of two patterns:

\begin{enumerate}
\item \emph{Phase-transition signature (supports Conjecture~3)}: Bimodality or a sharp inflection in the NS histogram at candidate \(\kappa^*\). Detection via three converging methods:
\begin{itemize}
\item \textbf{Hartigan's dip test} (Hartigan \& Hartigan, 1985): tests the null hypothesis of unimodality; $p < 0.05$ confirms bimodality.
\item \textbf{Kernel density estimation}: fit a Gaussian KDE to the NS distribution and identify inflection points $\partial^2 \hat{f}/ \partial \text{NS}^2 = 0$; the inflection closest to the $\tau(X)$ onset serves as the $\kappa^*$ candidate.
\item \textbf{Gaussian Mixture Model}: fit a two-component GMM to the NS values; if BIC favors two components over one, the crossover density between components estimates $\kappa^*$.
\end{itemize}
All three methods must agree within $\pm 0.05$ NS units to declare $\kappa^*$ empirically localized.

\item \emph{Tuning-parameter signature (weakens Conjecture~3)}: Continuous variation without concentration. If the NS distribution is approximately uniform or unimodal with no inflection, the thresholds are reclassified as continuous tuning parameters rather than phase transitions. This outcome does not falsify the three-regime model but significantly weakens its theoretical motivation.
\end{enumerate}

\noindent The upper threshold \(\kappa^{**}\) is estimated analogously: the NS value at which \(\tau(X)\) sharply declines or output quality collapses during the \(\alpha\)-sweep. The \(\kappa^{**}\) detection uses a changepoint analysis (PELT algorithm; Killick, Fearnhead, \& Eckley, 2012) on the time series \(\{\tau(X_i)\}_{i=1}^N\) ordered by increasing NS, identifying the point beyond which \(\beta_1\) persistence degrades catastrophically.

\subsubsection{5.4 Harmonic Maps and Cross-Fiber
Transport}\label{harmonic-maps-and-cross-fiber-transport}

The natural optimality criterion for the parallel transport map
\(\tau_{a \to b}: E_a \to E_b\) defined by connection \(\nabla\) is
Dirichlet energy minimization:

\[E(\tau) = \frac{1}{2} \int_{E_a} |d\tau|^2 \, dV_a\]

Maps minimizing \(E(\tau)\) are harmonic maps (Eells \& Sampson, 1964;
Jost, 2017). If the parallel transport map is harmonic, cross-domain
transport introduces no superfluous distortion---it is optimal in the
energy-cost sense.

\textbf{Open question for mathematical collaborators}: Under what
conditions is the parallel transport map harmonic? Does the
inseparability condition \(\text{NS} \geq \kappa\) impose a lower bound
\(E(\tau) \geq E_{\min}(\kappa)\) on Dirichlet energy? If so, does this
bound relate to the curvature norm \(\|F_\nabla\|^2\) via a known
inequality?

\subsubsection{5.5 Coordinate-Free Invariants: Persistent
Homology}\label{coordinate-free-invariants-persistent-homology}

The topology of the fiber \(E_b\) can be characterized without
presupposing manifold structure, through persistent homology
(Edelsbrunner \& Harer, 2010). Given neural representation point clouds
satisfying the inseparability signature, we construct Vietoris-Rips
complexes at multiple scales and extract Betti numbers: \(\beta_0\)
(connected components), \(\beta_1\) (1-dimensional loops), \(\beta_2\)
(2-dimensional voids).

\textbf{Topological signature contrast:}

{\def\LTcaptype{none} 
\begin{longtable}[]{@{}
  >{\raggedright\arraybackslash}p{(\linewidth - 4\tabcolsep) * \real{0.1774}}
  >{\raggedright\arraybackslash}p{(\linewidth - 4\tabcolsep) * \real{0.3387}}
  >{\raggedright\arraybackslash}p{(\linewidth - 4\tabcolsep) * \real{0.4839}}@{}}
\toprule\noalign{}
\begin{minipage}[b]{\linewidth}\raggedright
Hypothesis
\end{minipage} & \begin{minipage}[b]{\linewidth}\raggedright
\(\beta_0\) prediction
\end{minipage} & \begin{minipage}[b]{\linewidth}\raggedright
\(\beta_1, \beta_2\) prediction
\end{minipage} \\
\midrule\noalign{}
\endhead
\bottomrule\noalign{}
\endlastfoot
Rapid sequential processing & Two persistent independent components &
Trivial \\
Overlap Zone (inseparable) & Single connected component & Persistent
non-trivial loops and voids \\
\end{longtable}
}

\textbf{Structural tension index:}

\[\tau(X) = \frac{\beta_1 \text{ total persistence}}{\beta_0 \text{ total persistence}}\]

\textbf{Pathological mirror hypothesis}:
\(\tau(X_{\text{creative}}) > \tau(X_{\text{control}}) > \tau(X_{\text{pathological}})\).
\textbf{Falsification condition}: If \(\tau(X)\) for the creativity
group and the pathology group are statistically indistinguishable under
equivalence testing (TOST, $d < 0.2$), the theory's core distinction
collapses.

\textbf{Methodological note on equivalence testing.} The falsification condition above deliberately employs Two One-Sided Tests (TOST; Schuirmann, 1987; Lakens, 2017) rather than null-hypothesis significance testing (NHST). Standard NHST cannot confirm the null: failure to reject $H_0\!: \tau_{\text{creative}} = \tau_{\text{pathological}}$ may reflect low statistical power rather than genuine equivalence. TOST reverses the logic---the null hypothesis is that the two distributions \emph{differ} by at least the equivalence margin $\Delta$, and both one-sided tests must reject for equivalence to be declared:
\[
H_0^{-}\!: \mu_1 - \mu_2 \leq -\Delta \quad \text{and} \quad H_0^{+}\!: \mu_1 - \mu_2 \geq \Delta.
\]
Equivalence is declared only when both $H_0^{-}$ and $H_0^{+}$ are rejected at level $\alpha$, yielding a $(1-2\alpha)$ confidence interval for $\mu_1 - \mu_2$ that falls entirely within $(-\Delta, \Delta)$.

The choice of $d < 0.2$ (Cohen's $d$) as the equivalence margin rests on three considerations: (i)~Cohen's (1988) classification of $d = 0.2$ as the boundary of a ``small'' effect, below which differences lack practical significance; (ii)~the principle that a theoretical distinction claiming to separate creative cognition from pathological processing should manifest at least a small effect---if the topological signatures are indistinguishable at this threshold, the distinction carries no empirical content; (iii)~sensitivity analysis showing that tighter margins ($d < 0.1$) require prohibitively large samples ($N > 700$ per group) without proportional gain in theoretical discrimination.

\emph{Sample size derivation.} For a TOST procedure with equivalence margin $\Delta = 0.2$, power $1 - \beta = 0.80$, and $\alpha = 0.05$ (per one-sided test), the required sample size per group is:
\[
N = \frac{2(z_{1-\alpha} + z_{1-\beta})^2}{\Delta^2} \approx \frac{2(1.645 + 0.842)^2}{0.04} \approx 192.
\]
The experimental protocol in \S8.2 is powered accordingly. A pre-registered sensitivity analysis will also report results at $d < 0.15$ and $d < 0.3$ to assess robustness of the equivalence conclusion across reasonable margin choices.

\textbf{Bottleneck distance} \(W_\infty\) quantifies the decision: if
\(W_\infty\) between the overlap persistence diagram
\(D_{\text{overlap}}\) and any linear combination of single-modality
diagrams exceeds \(\epsilon\) (significantly above statistical noise),
with the difference driven primarily by non-trivial higher Betti numbers
(\(\beta_1, \beta_2 > 0\)), this mathematically demonstrates that the
system generates a topological structure irreducible to single-modality
processing. The bottleneck distance metric is preferable to Wasserstein
distance here because it captures the worst-case deviation in
persistence, which is the relevant quantity for demonstrating
\emph{qualitative} topological difference rather than average
distributional shift.

\subsubsection{5.6 Relation to Geometric Deep
Learning}\label{relation-to-geometric-deep-learning}

The framework maps onto the equivariance principles of the GDL program
(Bronstein et al., 2021). UOO can be restated in GDL vocabulary: a
\textbf{\(G\)-equivariant kernel} acting on the space of fiber bundle
sections over \(B\). Its equivariance means that regardless of how the
modality ``coordinate system'' varies, UOO's output representation
remains structurally invariant---the precise mathematical content of
``different modality pairs instantiate the same Overlap Zone
structure.''

Standard GDL handles intra-modal fiber structure (e.g., rotational
equivariance on images); UOO extends this with a cross-modal connection
on \(B\), giving geometric meaning to inter-modal representational
transport.

\textbf{Message passing as computational complement}: Bidirectional
semantic coupling (UOO condition ii) can be modeled as message passing
between propositional and non-propositional nodes (Gilmer et al., 2017).
Over-squashing---information loss through fixed-dimension bottlenecks in
deep GNNs (Alon \& Yahav, 2021)---is the Overlap Zone's natural
adversary: it is the graph-level mechanism that enforces contact
topology by constricting cross-modal information flow.
Anti-over-squashing techniques (graph rewiring, multi-scale
architectures, curvature-driven attention; Topping et al., 2022) can be
applied directly as computational complements to the fiber bundle
framework.

\subsubsection{5.7 Yang-Mills Three-Regime
Landscape}\label{yang-mills-three-regime-landscape}

The overlap/superimposition opposition can receive precise curvature
norm characterization. Let \(\nabla\) be a connection on the principal
bundle \(P(B, G)\); define the Yang-Mills action functional:

\[\|F_\nabla\|^2 = \int_B \text{tr}(F_\nabla \wedge *F_\nabla)\]

Three regimes correspond to curvature intervals:

\textbf{Regime I (Contact topology --- current AI):} No fiber bundle is
constructed---the system operates on disconnected modality-specific
manifolds. No connection, hence no curvature.

\textbf{Regime II (Overlap Zone --- creative cognition / UOO):} The
connection exists with bounded, non-zero curvature:
\(0 < \|F_\nabla\|^2 < C\). Non-zero curvature means cross-domain
mapping is not perfectly path-independent, but boundedness means the
system remains coherent. Non-trivial \(\beta_1\) features in persistent
homology are maintained.

\textbf{Regime III (Superimposition collapse --- psychopathology / UOO
failure):} The curvature norm exceeds \(C\) or diverges. The connection
develops singularities; fiber structure degenerates; originally distinct
representational dimensions become indistinguishable.

\begin{quote}
\textbf{Key correction}: Zero curvature (flat connection) corresponds to
an idealized limit---perfect cross-domain isomorphism. Real creative
overlap operates in the non-zero but bounded curvature regime. The flat
connection is the special case when the isomorphism conjecture
(Conjecture 2) holds perfectly, not the general requirement for the
Overlap Zone.
\end{quote}

\begin{quote}
\textbf{Editor's note}: This three-regime model has a precise analogy
with the structure of Yang-Mills theory---smooth configurations
vs.~singular instantons. Elevating this from analogy to rigorous
formalization requires precise specification of \(B\), \(G\), and
\(C\)---these remain priority open problems.
\end{quote}

\textbf{On the boundary constant \(C\).} The critical curvature threshold \(C\) that separates Regime~II (overlap) from Regime~III (collapse) is an open problem whose resolution admits at least three candidate paths:

\begin{enumerate}
\item \textbf{Empirical estimation from the PoC.} During the \(\alpha\)-sweep of the synthetic data PoC (\S6.3), the empirical distribution of \(\|F_\nabla\|^2\) will be recorded. The value at which \(\tau(X)\) undergoes a sharp downward transition---detected via the same changepoint analysis used for \(\kappa^{**}\)---provides a data-driven estimate of \(C\). This approach is model-agnostic but structure-group-dependent: different choices of \(G\) will yield different curvature scales.

\item \textbf{Derivation from the Casimir invariant.} For a compact semisimple Lie group \(G\), the quadratic Casimir operator \(C_2(G) = \sum_a T^a T^a\) (where \(\{T^a\}\) are generators of the Lie algebra \(\mathfrak{g}\) in the adjoint representation) provides a natural curvature scale intrinsic to the structure group. Specifically, for \(G = \mathrm{U}(1)\), \(C_2 = 1\); for \(G = \mathrm{SU}(2)\), \(C_2 = 3/4\) in the fundamental representation. The boundary constant may take the form \(C = \gamma \cdot C_2(G) \cdot \mathrm{vol}(B)\) for some dimensionless constant \(\gamma\) determined by the bundle topology (e.g., Chern number). This path connects the regime boundary directly to the structure group's representation theory and is the preferred theoretical resolution.

\item \textbf{Topological constraint from characteristic classes.} The Chern-Weil theorem relates \(\int_B \mathrm{tr}(F_\nabla \wedge F_\nabla)\) to the second Chern number \(c_2(P)\), a topological invariant. For bundles of fixed topology, this imposes a lower bound on \(\|F_\nabla\|^2\) that cannot be reduced by gauge transformation. If the overlap regime requires specific topological types (e.g., non-trivial \(c_2\)), this imposes a floor on \(C\) below which the regime cannot exist.
\end{enumerate}

\noindent The three paths are not mutually exclusive: empirical estimation (1) provides the target value; Casimir derivation (2) provides theoretical grounding; characteristic class constraints (3) determine whether the target is topologically achievable. We recommend that the first PoC report include an empirical \(C\) estimate alongside a comparison with the Casimir prediction for the chosen structure group.

\subsubsection{5.8 The Cruciform Structure in Fiber Bundle
Language}\label{the-cruciform-structure-in-fiber-bundle-language}

\begin{figure}[htbp]
\centering
\includegraphics[width=\textwidth]{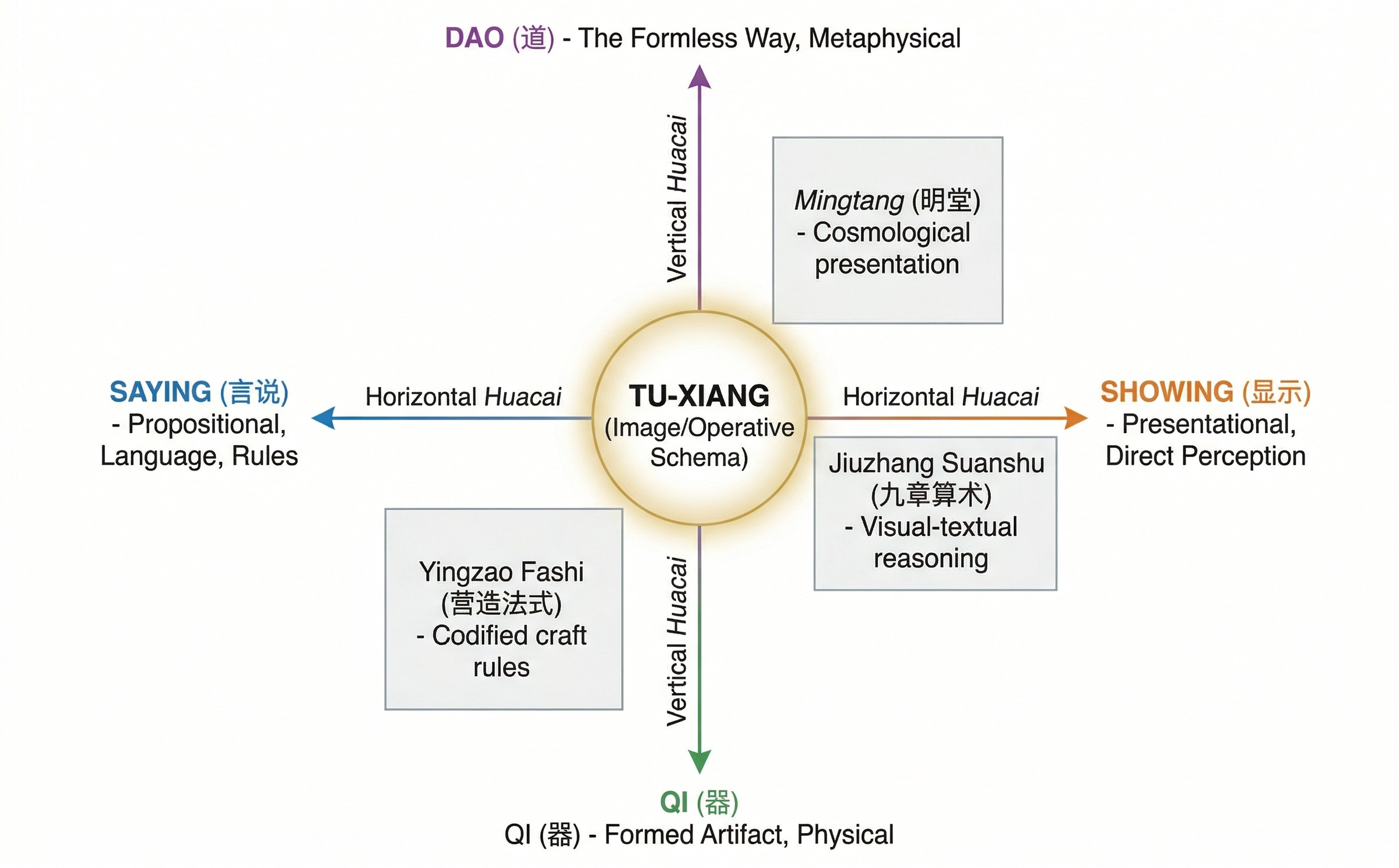}
\caption{\textbf{The Cruciform Structure.} \emph{Xiang} (tu-xiang, operative schema) occupies the intersection of two philosophical axes: the vertical \emph{dao}/\emph{qi} axis (metaphysical/physical) and the horizontal saying/showing axis (propositional/presentational). \emph{Xiang} simultaneously executes dual \emph{huacai} (transformation-and-cutting) along both axes. Four materials from the Chinese craft tradition each occupy a precise coordinate within this structure.}
\end{figure}

The philosophical cruciform structure of §3.3 admits a precise mapping
into the fiber bundle framework:

\begin{itemize}
\item
  \textbf{Vertical axis (\emph{dào}$\leftrightarrow$\emph{qì})} maps to the projection
  \(\pi: E \to B\)---the fiber over each point \(b \in B\) contains all
  possible representational states, from the most abstract (close to
  \emph{dào}) to the most concrete (close to \emph{qì}). The fiber
  direction is the \emph{dào}$\leftrightarrow$\emph{qì} direction.
\item
  \textbf{Horizontal axis (saying$\leftrightarrow$showing)} maps to the internal
  structure of each fiber \(E_b\)---the two representational dimensions
  (propositional and non-propositional) that the Overlap Zone requires
  to be inseparable. The non-separability entropy NS quantifies how
  deeply intertwined these dimensions are within a given fiber.
\item
  \textbf{\emph{Xiàng} as stable section} maps to a section
  \(s: B \to E\) satisfying \(\text{NS}(s(b)) \geq \kappa^*\) for all
  \(b\)---a cross-domain assignment that maintains inseparability across
  modality configurations. The isomorphism conjecture (§5.1) asserts
  that such sections exist and are structurally invariant.
\item
  \textbf{\emph{Chuànghuà} (creative transformation)} maps to the Neural
  ODE trajectory in which \(\beta_1\) features dynamically emerge: the
  continuous path \(z(0) \to z(T)\) during which the representation
  transitions from trivial (\(\beta_1 = 0\)) to non-trivial
  (\(\beta_1 > 0\)) topology. This is the computational instantiation of
  the creative event at the saying/showing intersection.
\item
  \textbf{\emph{Huàcái} (institutional cutting)} maps to the trained
  parameters \(\theta^*\) that encode the learned connection
  \(\nabla\)---the institutionalized operator that can be repeatedly
  applied to new inputs. Every inference pass through the trained UOO
  re-enacts, in compressed form, the creative dynamics discovered during
  training.
\item
  \textbf{Superimposition collapse} maps to section degeneration:
  \(s(b) \to\) singularity, NS \(\to \kappa^{**}\) or beyond,
  \(\beta_1 \to 0\)---the fiber structure collapses and dimensional
  distinction is destroyed.
\end{itemize}

This mapping demonstrates that the philosophical cruciform structure is
not an analogy for the mathematics---it is the same structure expressed
in two different formal languages. The fiber bundle is the
mathematician's cruciform; the cruciform is the philosopher's fiber
bundle. The UOO program seeks to build the engineer's implementation of
both.

\begin{center}\rule{0.5\linewidth}{0.5pt}\end{center}

\subsection{6. Computational Pillar: From Theory to
Implementation}\label{computational-pillar-from-theory-to-implementation}

\subsubsection{6.1 Why Current Architectures Suppress the Overlap
Zone}\label{why-current-architectures-suppress-the-overlap-zone}

\begin{quote}
\textbf{Core claim}: This limitation belongs to the level of inductive
biases and training objectives, not architectural impossibility.
\end{quote}

In the tensor product space \(V \otimes W\), non-separable tensors form
an open dense set---nearly all tensors are non-separable. A randomly
initialized neural network acting on \(V \otimes W\) almost surely
produces non-separable intermediate representations. Non-separability is
\emph{generic}, not rare.

Why, then, do trained systems fail to maintain non-separability? Because
their training objectives systematically push representations toward the
separable region:

\begin{enumerate}
\def\labelenumi{\arabic{enumi}.}
\tightlist
\item
  \textbf{Contrastive learning bias}: The global minimum of cosine
  similarity loss corresponds to fully separable embedding pairs.
\item
  \textbf{Attention structure bias}: Modality traceability reintroduces
  separability at every layer---each token retains its modal origin
  label.
\item
  \textbf{Denoising objective bias}: Diffusion architectures \emph{can}
  represent entangled states; the denoising objective drives the system
  out. Modified objectives (representation quality metrics at
  intermediate steps) can train diffusion models to dwell in entangled
  states.
\end{enumerate}

The question is therefore not ``how to build an architecture capable of
producing non-separable representations'' (nearly any architecture can)
but \textbf{``how to design training objectives that maintain and
reinforce non-separable representations.''} This reframing directly
motivates the topological regularization below.

\subsubsection{6.2 Three-Phase Architecture: Neural ODEs with
Topological
Regularization}\label{three-phase-architecture-neural-odes-with-topological-regularization}

\textbf{Phase 1: Bilinear entanglement (breaking contact topology).}
Replace simple concatenation \([x, y]\) with tensor product projection,
mapping propositional input \(x \in \mathbb{R}^{d_1}\) and
non-propositional input \(y \in \mathbb{R}^{d_2}\) through a learnable
nonlinear bilinear kernel onto a high-dimensional modality configuration
manifold:

\[z(0) = W_{\text{entangle}}(x \otimes y) \in \mathbb{R}^D, \quad D \gg \max(d_1, d_2)\]

The initial entangled state \(z(0)\) is the starting point for
continuous dynamics. The full tensor product
\(x \otimes y \in \mathbb{R}^{d_1 \times d_2}\) is computationally
expensive; practical implementations may use Tucker decomposition or
low-rank approximations to \(W_{\text{entangle}}\) while monitoring NS
to ensure the approximation does not inadvertently enforce separability.
Specifically, if the rank of the Tucker core falls below a critical
value, the approximation may project out precisely the non-separable
components that the architecture is designed to maintain.

\textbf{Phase 2: Continuous fiber bundle solver.} A parameterized neural
network \(f_\theta\) serves as the connection 1-form \(\nabla\) on the
fiber bundle, defining how representations continuously evolve along the
manifold:

\[\frac{dz(t)}{dt} = f_\theta(z(t), t)\]

Integration over \(t \in [0, T]\) naturally models the three-phase
dynamics observed in cognitive science: fusion phase (\(t \to 0\), high
geometric stress), oscillation phase (\(t \to T/2\), overcoming
topological barriers), stabilization phase (\(t \to T\), reaching the
harmonic state---\emph{xiàng}). The ODE solver choice matters:
adaptive-step solvers (Dormand-Prince) allow the system to take finer
steps during high-curvature regions of representation space, naturally
allocating more computational resources to topologically challenging
transitions. Fixed-step solvers risk either insufficient resolution in
critical regions or unnecessary computation in smooth regions.

\textbf{Phase 3: Topological regularization loss.} Without explicit
constraints, deep networks readily collapse \(z(T)\) back into
low-dimensional separable representations to reduce reconstruction
error. The topological loss writes inseparability into the optimization
objective:

\[\mathcal{L}_{\text{topo}} = \sum_{c \in PD(\beta_0)} \text{Lifetime}(c)^2 - \lambda_{\text{topo}} \sum_{k \in \{1,2\}} \sum_{h \in PD(\beta_k)} \text{Lifetime}(h)^2\]

This penalizes \(\beta_0\) (preventing the point cloud from fragmenting
into separate modal clusters) and rewards \(\beta_1, \beta_2\) (forcing
the network to maintain cross-modal topological entanglement). The
persistent homology computation uses the differentiable topological
layer of Hofer et al.~(2017), with algorithmic complexity \(O(n^3)\) in
the number of points; for mini-batch training, we compute persistence
diagrams on batches of \(z(T)\) representations. This \(O(n^3)\) cost is
comparable to attention computation and does not represent a
computational bottleneck for typical batch sizes.

\textbf{Gradient stability of \(\mathcal{L}_{\text{topo}}\).} A known limitation of differentiable persistent homology is gradient instability near degenerate persistence points (birth \(\approx\) death). When a topological feature is close to appearing or disappearing, the gradient of its lifetime with respect to input coordinates can exhibit discontinuities or numerical blow-up, because persistence diagrams are piecewise-linear functions of the input filtration values with non-smooth transitions at simplex additions. Three mitigation strategies are built into the PoC implementation:

\begin{enumerate}
\item \textbf{Soft thresholding}: Exclude persistence features with lifetime below a minimum threshold \(\epsilon_{\min} = 10^{-4}\) from the loss computation, avoiding gradient contributions from near-degenerate points. This threshold is set well below the meaningful persistence scale (\(\sim 10^{-1}\)) and affects only noise-level features.
\item \textbf{Gradient norm monitoring}: Track \(\|\nabla_{z(T)} \mathcal{L}_{\text{topo}}\|_2\) per training step and flag batches where the topological gradient exceeds \(10\times\) the running median. If more than 5\% of batches are flagged over any 100-step window, reduce \(\alpha\) or increase \(\epsilon_{\min}\).
\item \textbf{Backend robustness}: The PoC will benchmark two TDA backends---\texttt{giotto-ph} (Burella P\'erez et al., 2021) and \texttt{Ripser++} (Zhang et al., 2020)---and report gradient agreement. Discrepancies exceeding \(10^{-3}\) in the gradient direction (cosine distance) would indicate numerical unreliability requiring backend-specific fixes.
\end{enumerate}

\noindent The PoC report will include a dedicated ``Gradient Health'' appendix documenting \(\|\nabla \mathcal{L}_{\text{topo}}\|_2\) trajectories, degenerate-point frequency, and backend comparison results. This transparency ensures that any topological training signal claimed is numerically trustworthy.

\textbf{Total loss:}

\[\mathcal{L}_{\text{Total}} = \mathcal{L}_{\text{Task}}(z(T), Y_{\text{target}}) + \alpha \mathcal{L}_{\text{topo}}\]

where \(\alpha\) controls the non-separability enforcement strength. The
dual-threshold model (§5.3) suggests that \(\alpha\) should be tuned to
maintain \(\kappa^* \leq \text{NS}(z(T)) < \kappa^{**}\)---too low and
the system reverts to contact topology; too high and it risks
computational superimposition collapse. In practice, \(\alpha\) can be
scheduled: beginning with a higher value to establish non-separable
representations, then gradually reducing to allow task-specific
fine-tuning while monitoring NS to ensure it remains within the overlap
regime.

\begin{figure}[htbp]
\centering
\includegraphics[width=\textwidth]{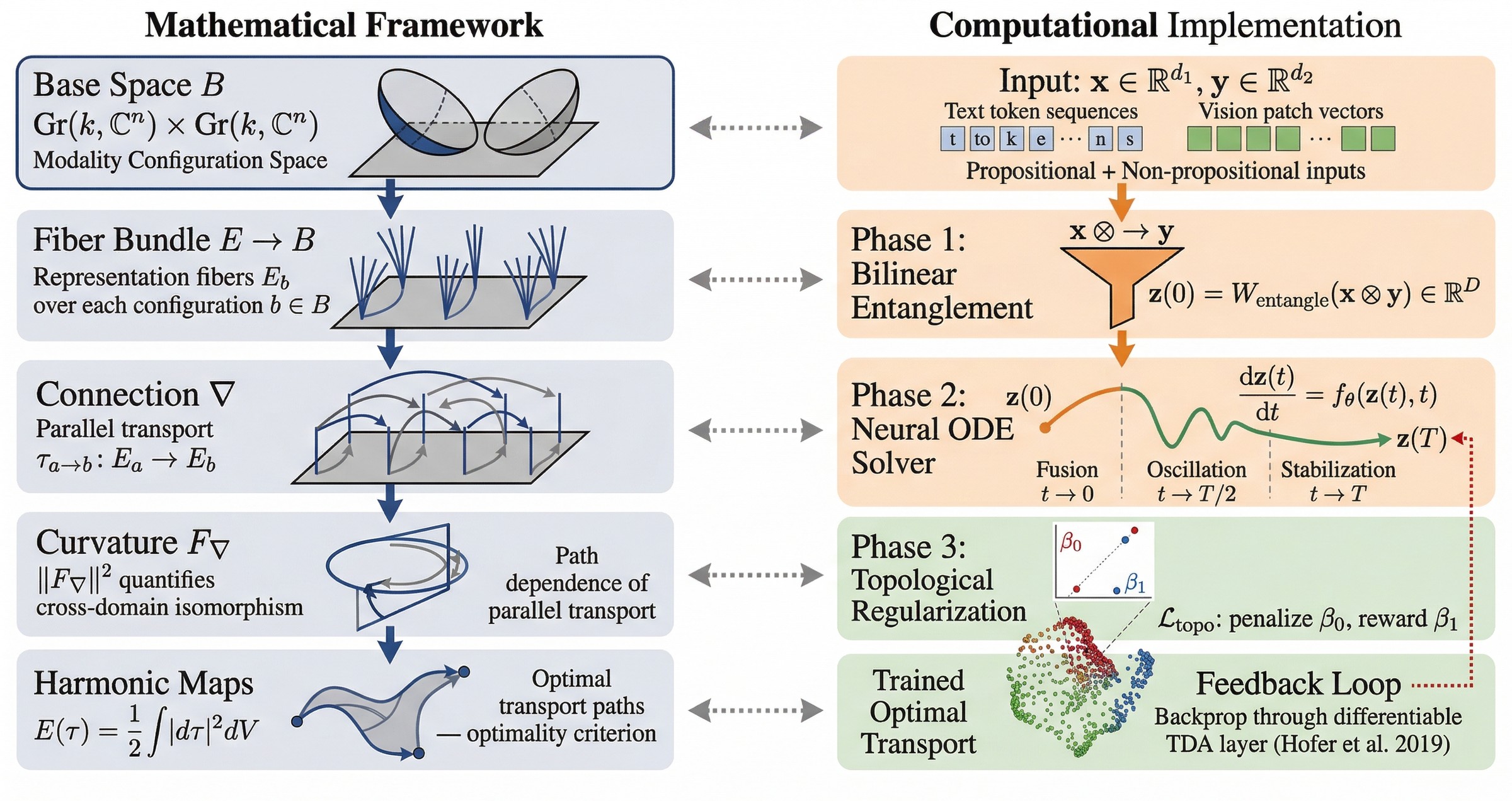}
\caption{\textbf{The UOO Computation Graph.} Left: mathematical framework (base space $B$, fibers $E_b$, connection $\nabla$, curvature $F_\nabla$, harmonic maps). Right: computational implementation (bilinear entanglement $\to$ Neural ODE $\to$ topological regularization). Correspondence arrows map each mathematical construct to its computational realization.}
\label{fig:uoo-computation}
\end{figure}

Figure~\ref{fig:uoo-computation} provides a side-by-side visualization of the
mathematical framework (fiber bundle, connection, curvature, harmonic
maps) and its computational implementation (bilinear entanglement,
Neural ODE, \(\mathcal{L}_{\text{topo}}\)), with explicit
correspondence arrows mapping each mathematical construct to its
computational realization.

\textbf{Scalability roadmap.} The \(O(n^3)\) cost of persistent homology
computation deserves explicit staging across the research program's
phases. In Phase 1 (synthetic data PoC, point clouds of 256--512
points), direct Vietoris-Rips computation via Ripser (Bauer, 2021) is
sufficient---\(O(n^3)\) at this scale is comparable to a single
attention layer and does not constitute a training bottleneck. In Phase
2 and beyond, as representation dimensionality and batch sizes increase,
two complementary acceleration strategies become relevant. First,
\textbf{landmark-based Witness Complexes} (de Silva \& Carlsson, 2004):
select \(m \ll n\) landmark points via max-min sampling and compute the
Witness Complex on these landmarks, reducing complexity to \(O(m^3)\)
with \(m\) typically 50--100 even for large point clouds---a strategy
validated in the Topological Autoencoder pipeline (Moor et al., 2020).
Second, \textbf{Distance-to-Measure (DTM) filtration} (Anai et al.,
2020): replace the Vietoris-Rips filtration with a DTM-based filtration
that is both robust to outliers and amenable to \(k\)-d tree
acceleration, achieving \(O(n \log n)\) complexity for the filtration
construction step. We emphasize an architectural design principle: UOO
need not replace all Transformer layers. In production-scale systems,
UOO can function as a \textbf{high-order cognitive plug-in}---inserted
after feature extraction and before final generation---dedicated
specifically to cross-modal structural reasoning, while standard
attention handles intra-modal processing. This modular design bounds the
topological computation cost to a fixed fraction of total inference.

\subsubsection{6.3 Synthetic Data Minimum Viable Proof of
Concept}\label{synthetic-data-minimum-viable-proof-of-concept}

Before accessing cognitive neuroscience data (requiring months to
years), synthetic experiments provide rapid low-cost
validation---results within weeks:

\begin{enumerate}
\def\labelenumi{\arabic{enumi}.}
\tightlist
\item
  \textbf{Construct two synthetic modalities} (e.g., two families of
  random graphs with distinct community structures, or two families of
  synthetic images with independent geometric regularity), each with
  independent structural regularities. The choice of synthetic
  modalities is deliberate: graph community structure and geometric
  regularity are independently manipulable, their entanglement can be
  precisely controlled, and the ground truth non-separability of
  task-relevant features is known by construction.
\item
  \textbf{Define a synthetic Overlap Zone task}: a classification or
  regression task requiring simultaneous use of both modalities'
  structure, with the two modalities' information entangled in a
  non-separable manner within the task structure. Crucially, the task
  must be designed so that no linear combination of single-modality
  features suffices for optimal performance---the task itself demands
  non-separable processing. This can be verified by establishing a
  theoretical performance ceiling for separable approaches.
\item
  \textbf{Comparatively train three conditions}: (a) CLIP-style
  contrastive alignment (separability-biased baseline); (b) Neural ODE
  with topological regularization (\(\mathcal{L}_{\text{topo}}\), non-separability-biased); (c) Neural ODE with \(\alpha = 0\) (no topological regularization---continuous-time dynamics only). All
  three conditions should be matched in parameter count and computational
  budget to ensure that performance differences reflect representational
  structure rather than capacity.

  \textbf{Rationale for the \(\alpha = 0\) ablation.} Condition~(c) is the critical ablation control that isolates the contribution of the topological regularization term from the benefits of continuous-time dynamics. The Neural ODE architecture itself (adaptive-step integration, continuous trajectory through representation space) may confer advantages over the discrete attention layers in the CLIP baseline, independent of any topological constraint. Without condition~(c), a positive result for~(b) versus~(a) would be ambiguous: it could reflect either the topological regularization or the continuous-time architecture or both. The three-condition design yields a \(2 \times 2\) factorial interpretation: \{discrete vs.\ continuous\} \(\times\) \{separable vs.\ non-separable objective\}, with condition~(c) occupying the \{continuous, separable\} cell. The expected ordering under the Overlap Zone hypothesis is: (b)~\(>\)~(c)~\(>\)~(a) on zero-shot transfer tasks requiring cross-modal structural reasoning, with (b)~\(>\)~(c) driven specifically by \(\beta_1\) persistence and (c)~\(>\)~(a) driven by trajectory smoothness.
\item
  \textbf{Measure NS and \(\beta_1\) persistence profiles} throughout
  training for both architectures, validating that the topological
  regularization achieves its intended effect on representational
  topology.
\item
  \textbf{Test zero-shot transfer on new synthetic modality pairs} that
  were not seen during training, measuring both task performance and
  topological profile preservation.
\item
  \textbf{Termination criterion}: If the non-separability bias does
  \emph{not} significantly outperform the separability bias on zero-shot
  transfer, terminate the computational program at minimal cost; if it
  does, proceed to Phase 1.5 and beyond
\end{enumerate}

\begin{center}\rule{0.5\linewidth}{0.5pt}\end{center}

\subsection{7. UOO Failure Modes and ANALOGY-MM
Benchmark}\label{uoo-failure-modes-and-analogy-mm-benchmark}

\subsubsection{7.1 Three Stress Conditions and Failure
Predictions}\label{three-stress-conditions-and-failure-predictions}

Section 4.4.4 established that LLM hallucination is not superimposition
collapse. For a UOO-trained system, however, a precise failure mode
prediction exists:

\textbf{Prediction}: A UOO system that successfully maintains
non-separable representations will, under the following stress
conditions, see its representations degrade from Regime II to Regime
III:

\begin{enumerate}
\def\labelenumi{\arabic{enumi}.}
\tightlist
\item
  \textbf{Regularization failure}: If \(\mathcal{L}_{\text{topo}}\)
  weight \(\alpha\) trends toward zero during training, or is disabled
  at inference, the system will collapse representations into low-rank
  separable states---essentially reverting to contact topology.
\item
  \textbf{Out-of-distribution input}: When inputs far exceed the
  training distribution, the learned connection may fail, producing
  trajectories that exit the bounded curvature regime in representation
  space.
\item
  \textbf{Excessive entanglement}: If the training objective
  over-rewards non-separability without upper-bound constraints (i.e.,
  lacking a computational implementation of \(\kappa^{**}\)), the system
  may enter a highly entangled but semantically fragmented
  state---computational superimposition collapse.
\end{enumerate}

\textbf{Testing protocol}: On the UOO prototype system (after synthetic
data PoC), monitor NS and \(\beta_1\) total persistence under each
stress condition, verifying whether representational degradation is
accompanied by a sharp drop in \(\beta_1\) persistence and correlated
output quality degradation. The correlation between topological
degradation and output quality degradation would provide direct evidence
for the causal role of non-separable structure.

\textbf{RLHF and the separability restoring force.} A subtlety arises for systems trained with reinforcement learning from human feedback (RLHF; Christiano et al., 2017; Ouyang et al., 2022). One might hypothesize that RLHF---by optimizing for human-preferred outputs that may implicitly reward creative, cross-modal reasoning---could push representations toward the overlap regime even without explicit topological regularization. The framework predicts the opposite: RLHF acts as a \emph{restoring force toward contact topology}, not away from it.

The mechanism is as follows. The base training objective (contrastive, cross-attention, or denoising) encodes a separability bias into the loss landscape geometry (\S6.1). RLHF operates as a reward-weighted fine-tuning step that adjusts the policy within the representational manifold already shaped by this base objective. Preference optimization can modify the \emph{trajectory} through representation space but cannot restructure the \emph{topology} of the loss landscape itself. If a non-separable representation emerges transiently during preference optimization, the base objective's gradient field constitutes a persistent restoring force that pushes the system back toward separable states.

Formally, let \(\mathcal{L}_{\text{base}}\) denote the base training loss and \(\mathcal{L}_{\text{RLHF}}\) the reward model loss. The combined gradient field is:
\[
\nabla_\theta \mathcal{L} = \nabla_\theta \mathcal{L}_{\text{base}} + \beta \nabla_\theta \mathcal{L}_{\text{RLHF}}
\]
where \(\beta\) is the KL penalty coefficient. For any representation \(z\) with \(\text{NS}(z) > 0\), the base gradient component \(\nabla_\theta \mathcal{L}_{\text{base}}\) points toward lower NS (separable states), while the RLHF component may transiently point toward higher NS if the reward model favors non-separable outputs. Stable non-separability requires \(\beta \|\nabla \mathcal{L}_{\text{RLHF}}\| > \|\nabla \mathcal{L}_{\text{base}}\|\) \emph{persistently}---an unlikely equilibrium given that \(\beta\) is typically small and the KL penalty itself constrains deviation from the base policy.

\emph{Escape velocity prediction}: A UOO-augmented system with explicit \(\mathcal{L}_{\text{topo}}\) can achieve stable non-separability because the topological term directly counteracts the separability restoring force. A system relying solely on RLHF to induce overlap will exhibit intermittent, unstable non-separability that degrades under distribution shift---a testable distinction in the ANALOGY-MM benchmark.

\subsubsection{7.2 ANALOGY-MM: Cross-Modal Analogy
Benchmark}\label{analogy-mm-cross-modal-analogy-benchmark}

\textbf{Benchmark design.} Present a model with a cross-modal analogy
problem: ``Image A is to Image B as Concept C is to ? Concept,'' where
A:B embodies a structural relationship (spatial transformation,
containment, quantity change, attribute change), requiring the model to
complete the structural mapping in the linguistic domain.

\textbf{Concrete examples:} - \emph{Spatial}: A shows a square inside a
circle, B shows a triangle inside a circle → C = ``cat inside box'' → ?
= ``dog inside box'' (not ``dog outside box'') - \emph{Quantity}: A
shows three apples, B shows six apples → C = ``quiet room'' → ? =
``noisy room'' (doubling of intensity, not merely ``another room'') -
\emph{Attribute}: A shows a red circle, B shows a blue circle → C =
``hot coffee'' → ? = ``cold coffee'' (attribute inversion, not ``hot
tea'')

\textbf{Core prediction.} The \textbf{error type ratio} (ETR = errors
where content is correct but relational structure is wrong / total
errors) for all three architecture classes under cross-modal conditions
should be significantly higher than two single-modal baselines
(visual-only analogy, language-only analogy). This error type
shift---not lower scores per se, but errors systematically biased toward
``content acquired but structural mapping failed''---is the specific
signature of contact topology.

\textbf{Decision logic:}

{\def\LTcaptype{none} 
\begin{longtable}[]{@{}
  >{\raggedright\arraybackslash}p{(\linewidth - 2\tabcolsep) * \real{0.4359}}
  >{\raggedright\arraybackslash}p{(\linewidth - 2\tabcolsep) * \real{0.5641}}@{}}
\toprule\noalign{}
\begin{minipage}[b]{\linewidth}\raggedright
Outcome pattern
\end{minipage} & \begin{minipage}[b]{\linewidth}\raggedright
Implication for theory
\end{minipage} \\
\midrule\noalign{}
\endhead
\bottomrule\noalign{}
\endlastfoot
Cross-modal ETR significantly \textgreater{} single-modal ETR & Supports
contact topology diagnosis \\
No ETR difference across conditions & Opposes reality of topological
type distinction \\
Some architecture shows no ETR shift on cross-modal & That architecture
may have partially achieved overlap topology \\
\end{longtable}
}

\begin{figure}[htbp]
\centering
\includegraphics[width=\textwidth]{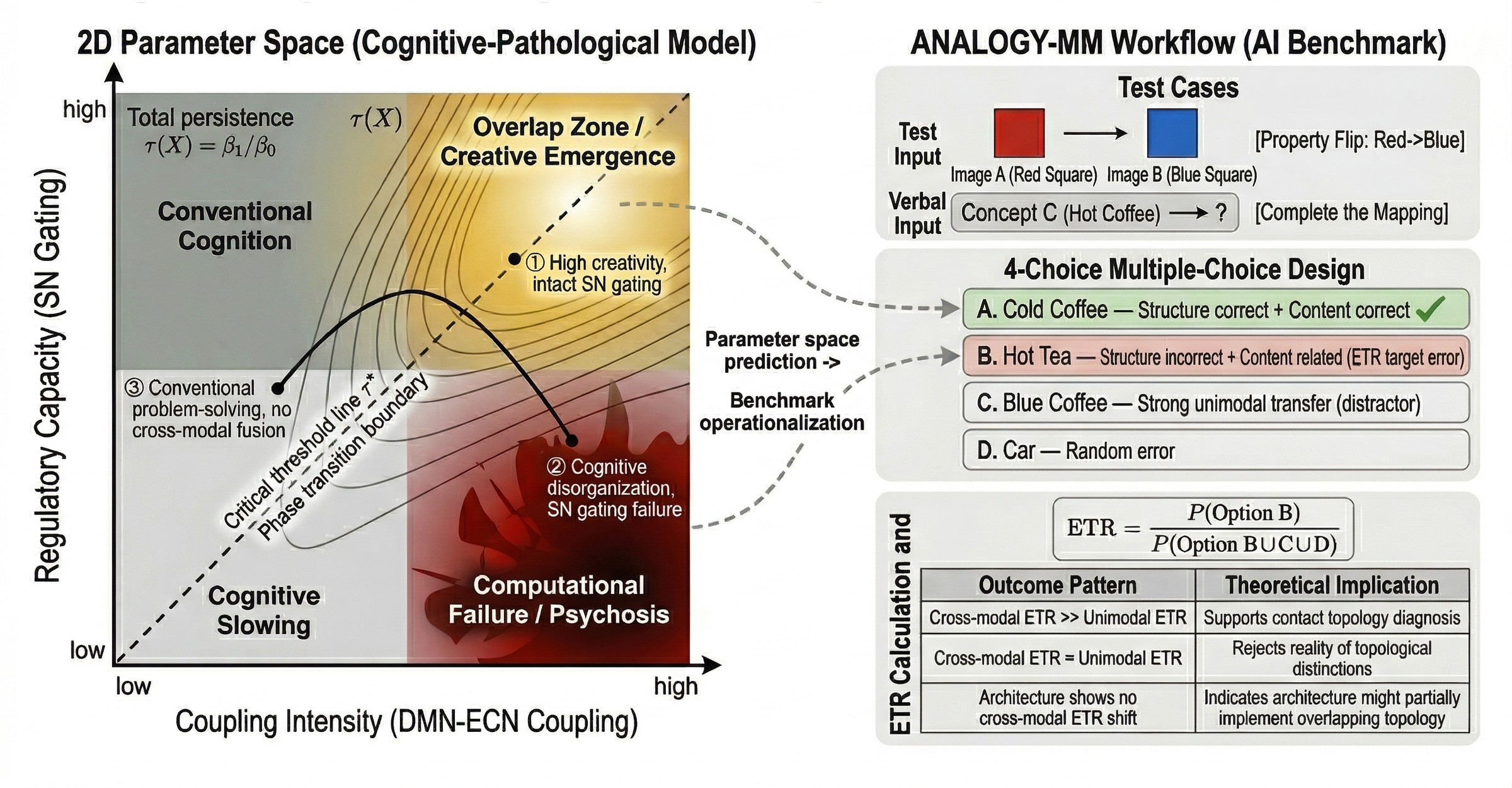}
\caption{\textbf{Cognitive-Pathological Parameter Space and ANALOGY-MM Benchmark.} Left: two-dimensional parameter space (coupling intensity $\times$ regulatory capacity) with four quadrants---Overlap Zone (high coupling, high regulation), Superimposition Collapse (high coupling, low regulation), Conventional Cognition (low coupling, high regulation), and Cognitive Slowing (low coupling, low regulation). $\tau(X)$ contour lines indicate predicted structural tension. Right: ANALOGY-MM cross-modal analogy workflow with 2$\times$2 forced-choice design and ETR computation.}
\label{fig:parameter-space}
\end{figure}

Figure~\ref{fig:parameter-space} juxtaposes the cognitive-pathological
two-dimensional parameter space (coupling intensity $\times$ regulatory
capacity, with \(\tau(X)\) contour lines) and the ANALOGY-MM evaluation
workflow, illustrating how the theoretical framework maps to an
immediately deployable computational diagnostic.

\subsubsection{7.3 Scalable Evaluation Protocol for
ANALOGY-MM}\label{scalable-evaluation-protocol-for-analogy-mm}

The ETR metric as described above requires classifying each error as
``content correct but structure wrong'' versus other error types. For
large-scale automated evaluation, we propose a \textbf{forced-choice
multiple-choice format} with a 2×2 factorial option design:

{\def\LTcaptype{none} 
\begin{longtable}[]{@{}
  >{\raggedright\arraybackslash}p{(\linewidth - 4\tabcolsep) * \real{0.3333}}
  >{\raggedright\arraybackslash}p{(\linewidth - 4\tabcolsep) * \real{0.3333}}
  >{\raggedright\arraybackslash}p{(\linewidth - 4\tabcolsep) * \real{0.3333}}@{}}
\toprule\noalign{}
\begin{minipage}[b]{\linewidth}\raggedright
\end{minipage} & \begin{minipage}[b]{\linewidth}\raggedright
Structure correct
\end{minipage} & \begin{minipage}[b]{\linewidth}\raggedright
Structure wrong
\end{minipage} \\
\midrule\noalign{}
\endhead
\bottomrule\noalign{}
\endlastfoot
\textbf{Content domain correct} & \textbf{A} (correct answer) &
\textbf{B} (ETR target error) \\
\textbf{Content domain wrong} & \textbf{C} (structure transfer, wrong
domain) & \textbf{D} (random / double error) \\
\end{longtable}
}

For each test item, the four options are constructed as follows.
\textbf{Option A} (correct): applies the source analogy's relational
structure to the target domain (e.g., red→blue applied to hot coffee →
cold coffee). \textbf{Option B} (ETR target): selects a semantically
related concept from the target domain's associative neighborhood
(cosine similarity \textgreater{} 0.7 in embedding space) but with the
wrong relational mapping (e.g., hot coffee → hot tea---staying within
the ``hot beverage'' semantic cluster without applying attribute
inversion). \textbf{Option C} (single-modal interference): correctly
captures the structural transformation but literally transfers the
source modality's surface feature into the target domain (e.g., hot
coffee → blue coffee---importing the visual attribute ``blue'' rather
than abstracting ``attribute inversion''). \textbf{Option D} (random):
an unrelated concept from a distant semantic domain (cosine similarity
\textless{} 0.2).

The automated ETR computation becomes:

\[\text{ETR} = \frac{P(\text{B})}{P(\text{B}) + P(\text{C}) + P(\text{D})}\]

i.e., among all incorrect responses, the proportion that are
``content-related but structurally wrong.''

\textbf{Open-generation calibration subset.} The forced-choice format
may underestimate contact topology's limitations by excluding
\textbf{overflow errors}---responses that fall outside all four option
categories (refusal to answer, circular repetition of input, generation
of unrelated novel concepts). To calibrate, a randomly sampled subset of
\textasciitilde100 items is additionally administered in open-generation
format, with three trained annotators independently coding each response
into category A/B/C/D/E (E = overflow). The calibration standard is
Cohen's \(\kappa \geq 0.80\) inter-annotator agreement and Pearson
\(r \geq 0.85\) between forced-choice \(\text{ETR}_{\text{mc}}\) and
open-generation \(\text{ETR}_{\text{open}}\). The overflow error rate
\(\text{OER} = P(\text{E}) / N_{\text{total}}\) serves as a
supplementary diagnostic: cross-modal OER should exceed single-modal OER
if contact topology genuinely constrains cross-modal reasoning.

\textbf{Domain coverage.} The benchmark spans five cross-modal domain
pairs (visual-spatial → language-conceptual: attribute inversion,
containment, quantity-intensity mapping; auditory-rhythmic →
language-prosodic: temporal structure; tactile-textural →
visual-textural: surface feature mapping), with three difficulty levels
per domain following Gentner's (1983) systematicity hierarchy (surface
mapping, relational mapping, system mapping), yielding approximately 200
items.

\subsubsection{7.4 META-TOP: From Structural Mapping to Dynamical
Topological
Isomorphism}\label{meta-top-from-structural-mapping-to-dynamical-topological-isomorphism}

ANALOGY-MM (§7.2--7.3) tests whether current architectures can perform
cross-modal \emph{structural mapping}---applying a relational
transformation from one modality domain to another. This is a necessary
but not sufficient condition for Overlap Zone processing. A deeper test
asks whether a system can detect \textbf{dynamical topological
isomorphism}: can it recognize that a deep cultural metaphor and a
physical dynamical system share the same underlying topological
evolution, even when they share no surface features?

We propose META-TOP (Metaphor-Topology) as a three-tier extension of
ANALOGY-MM, designed to test progressively deeper levels of cross-modal
topological understanding:

\textbf{Tier L1: ANALOGY-MM} (§7.2--7.3). Cross-modal structural mapping
with ETR metric. Immediately deployable on existing architectures.
Tests: attribute inversion, containment, quantity mapping across
modality pairs.

\textbf{Tier L2: META-TOP Simple.} Cross-modal dynamical pattern
recognition using synthetic dynamical systems paired with simplified
metaphorical text. The key metric shifts from ETR (binary error
classification) to \textbf{TSAS (Topological Structural Alignment
Score)}: the normalized bottleneck distance \(W_\infty\) between the
persistence diagrams extracted from the model's latent-space
trajectories when processing the text input versus the dynamical system
visualization. Lower TSAS indicates higher topological alignment. Human
expert rankings provide calibration via Kendall's \(\tau\) correlation.
\textbf{Prerequisite}: Phase 1 synthetic data PoC must demonstrate that
\(\mathcal{L}_{\text{topo}}\) produces detectable topological
differences.

\textbf{Tier L3: META-TOP Full.} Complete cross-civilizational
topological isomorphism testing. Seven topological archetypes---each
grounded in at least three independent civilizational traditions---are
paired with unlabeled physical dynamical system visualizations. The
archetypes include: (1) irreversible entropy / chaos-genesis; (2)
non-orientability / void-plenum; (3) critical phase transition; (4)
liminal passage / bifurcation; (5) topological collapse (the
computational mirror of superimposition collapse); (6) spiral ascent /
transformative refinement; (7) \emph{chuànghuà} (creative
transformation)---the dynamic emergence of \(\beta_1\) topology itself,
tested using texts from the \emph{Zhuangzi} (Cook Ding), the
\emph{Yijing} (hexagram generation), and cross-civilizational parallels.
\textbf{Prerequisite}: UOO prototype system + cross-cultural
psychometric validation of test items.

The critical innovation of META-TOP relative to existing multimodal
benchmarks is that Tier L3 Archetype 7 (\emph{chuànghuà}) tests not the
recognition of existing topological features but the model's ability to
detect \textbf{the dynamic generation process of topology itself}---the
transition from topologically trivial to topologically non-trivial
representation, which is the computational essence of the
creative-transformative zone.

\begin{figure}[htbp]
\centering
\includegraphics[width=\textwidth]{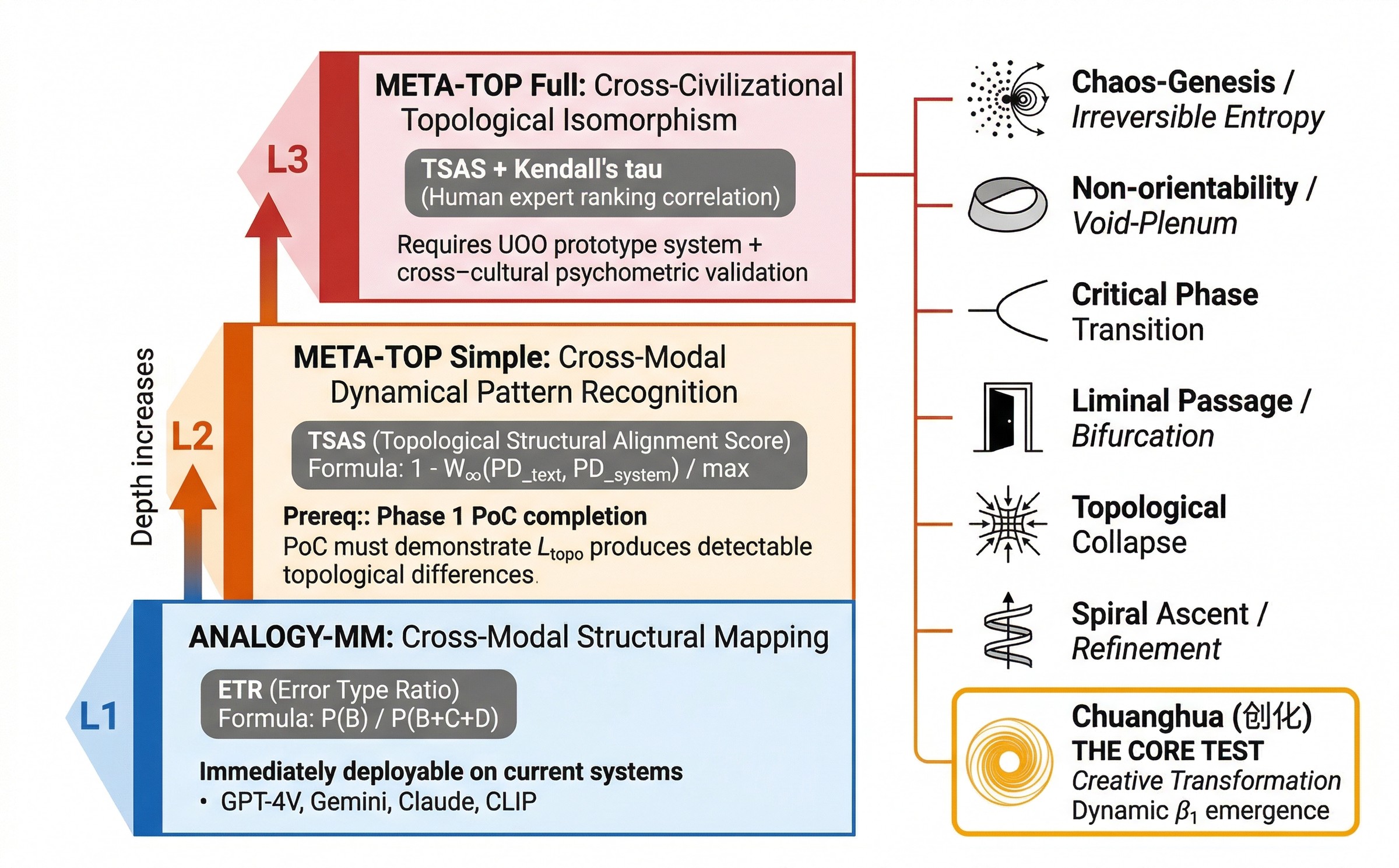}
\caption{\textbf{META-TOP --- Three-Tier Benchmark System.} Left: progressive three-tier structure testing increasingly deep cross-modal topological understanding. Tier L1 (ANALOGY-MM) tests structural mapping with ETR. Tier L2 (META-TOP Simple) tests dynamical pattern recognition with TSAS. Tier L3 (META-TOP Full) tests cross-civilizational topological isomorphism across seven archetypes. Right: the seven topological archetypes, with Archetype 7 (\emph{chuanghua}, creative transformation) as the core test---detecting the dynamic \emph{emergence} of topology itself.}
\end{figure}

\begin{center}\rule{0.5\linewidth}{0.5pt}\end{center}

\subsection{8. Research Agenda: Phased Experimental
Plan}\label{research-agenda-phased-experimental-plan}

\subsubsection{8.1 Sequential Roadmap}\label{sequential-roadmap}

We propose a strictly sequential plan, each phase contingent on positive
results from the preceding phase:

{\def\LTcaptype{none} 
\begin{longtable}[]{@{}
  >{\raggedright\arraybackslash}p{(\linewidth - 6\tabcolsep) * \real{0.1522}}
  >{\raggedright\arraybackslash}p{(\linewidth - 6\tabcolsep) * \real{0.1522}}
  >{\raggedright\arraybackslash}p{(\linewidth - 6\tabcolsep) * \real{0.2174}}
  >{\raggedright\arraybackslash}p{(\linewidth - 6\tabcolsep) * \real{0.4783}}@{}}
\toprule\noalign{}
\begin{minipage}[b]{\linewidth}\raggedright
Phase
\end{minipage} & \begin{minipage}[b]{\linewidth}\raggedright
Scope
\end{minipage} & \begin{minipage}[b]{\linewidth}\raggedright
Timeline
\end{minipage} & \begin{minipage}[b]{\linewidth}\raggedright
Termination criterion
\end{minipage} \\
\midrule\noalign{}
\endhead
\bottomrule\noalign{}
\endlastfoot
Phase 0 & White paper revision; candidate definitions for \(B\), \(G\),
\(E_b\); pathological mirror integration & 1--2 months & --- \\
Phase 1 & Synthetic data PoC + NS validation + synthetic TDA pipeline &
1--3 months & Separable bias $\geq$ non-separable on zero-shot transfer →
terminate \\
Phase 1.5 & TDA pipeline pre-validation on public fMRI creativity
datasets & 1--2 months & TDA metrics fail to distinguish high/low
creativity → redesign \\
Phase 2 & Behavioral experiment 1 + ANALOGY-MM first round & 3--6 months
& No cross-dimension semantic propagation → reassess framework \\
Phase 3 & Neuroimaging experiment 2 + dimensional pathological mirror
test & 6--12 months & No non-decomposable emergent component → fiber
bundle needs alternative \\
Phase 4 & Cross-domain isomorphism (Expts 3--4) + causal TMS & 12--24
months & Topological signatures differ across modality pairs →
isomorphism falsified \\
\end{longtable}
}

\subsubsection{8.2 Experiment 1: Behavioral Validation (Dimensional
Design)}\label{experiment-1-behavioral-validation-dimensional-design}

\textbf{Paradigm}: Selectively perturb one dimension in a creative
cross-modal task; measure whether the other dimension exhibits
semantically correlated synchronous changes.

{\def\LTcaptype{none} 
\begin{longtable}[]{@{}
  >{\raggedright\arraybackslash}p{(\linewidth - 2\tabcolsep) * \real{0.5862}}
  >{\raggedright\arraybackslash}p{(\linewidth - 2\tabcolsep) * \real{0.4138}}@{}}
\toprule\noalign{}
\begin{minipage}[b]{\linewidth}\raggedright
Design dimension
\end{minipage} & \begin{minipage}[b]{\linewidth}\raggedright
Enhancement
\end{minipage} \\
\midrule\noalign{}
\endhead
\bottomrule\noalign{}
\endlastfoot
\textbf{Participant design} & \textbf{Dimensional individual-differences
design}: recruit 150--200 healthy participants, continuously measuring
creativity (AUT originality + DAT + CAQ) and schizotypy (O-LIFE scale),
using continuous regression instead of between-group ANOVA \\
\textbf{Primary metric} & Automated semantic distance (SemDis) as
primary quantification; human ratings as validation only \\
\textbf{Demand characteristic control} & Low-level perturbation control
condition (non-semantic perturbation) to exclude confounds \\
\textbf{Temporal resolution} & Continuous trajectory tracking (mouse
trajectory/force-sensitive input) throughout 5-second response window \\
\textbf{Sequential processing control} & 10Hz forced alternation control
group: if inseparability signature collapses under high-frequency
alternation, this evidences the Overlap Zone's dependence on temporal
co-presence \\
\textbf{Network framework} & DMN/ECN/SN tripartite analysis; SN
interpreted as coupling regulator (§4.2) \\
\textbf{Dynamic analysis} & Data-driven state discovery (BIC/AIC model
selection) replacing presupposed three-stage templates \\
\textbf{Theoretical adjudication} & Explicitly designed to distinguish
``optimal coupling'' model (graded) from ``non-separable processing
state'' model (threshold) (§4.2) \\
\textbf{Pathological mirror test} & Creativity $\leftrightarrow$ \(\beta_1\) persistence
(positive correlation); schizotypy cognitive disorganization $\leftrightarrow$
cross-trial \(\beta_1\) variance (topological instability, positive
correlation) \\
\end{longtable}
}

\textbf{Falsification criterion}: If selective perturbation produces
\emph{only} performance impairment (dual-task interference) without
semantically correlated changes, the Overlap Zone hypothesis is
falsified.

\subsubsection{8.3 Experiment 2: Neural Structure Analysis (Layered
Decision
Design)}\label{experiment-2-neural-structure-analysis-layered-decision-design}

Extract multi-voxel activation patterns via fMRI, combined with MEG's
millisecond temporal resolution, to identify emergent components not
decomposable as linear combinations of single-modality patterns.

\textbf{Layered decision design to exclude ``rapid sequential
processing'' alternative hypothesis:}

{\def\LTcaptype{none} 
\begin{longtable}[]{@{}
  >{\raggedright\arraybackslash}p{(\linewidth - 4\tabcolsep) * \real{0.2500}}
  >{\raggedright\arraybackslash}p{(\linewidth - 4\tabcolsep) * \real{0.2100}}
  >{\raggedright\arraybackslash}p{(\linewidth - 4\tabcolsep) * \real{0.5400}}@{}}
\toprule\noalign{}
\begin{minipage}[b]{\linewidth}\raggedright
Layer
\end{minipage} & \begin{minipage}[b]{\linewidth}\raggedright
Method
\end{minipage} & \begin{minipage}[b]{\linewidth}\raggedright
Decision criterion
\end{minipage} \\
\midrule\noalign{}
\endhead
\bottomrule\noalign{}
\endlastfoot
\textbf{Behavioral} & Continuous trajectory tracking & Rapid switching →
discrete stepwise polyline; Overlap Zone → smooth global topological
deformation \\
\textbf{Electrophysiological} & Zero-lag phase locking (MEG) & DMN and
ECN must simultaneously fire Gamma bursts within the same low-frequency
Theta phase slot---sharing a metronome, not alternating \\
\textbf{State space} & HMM microstate analysis & Must reveal a third
dynamical attractor state whose Betti number characteristics cannot be
reconstructed as linear superposition of ``pure propositional'' and
``pure imagistic'' states \\
\textbf{Causal (information-theoretic)} & Bidirectional transfer entropy
& Perturbation must trigger instantaneous nonlinear global network
reorganization (not sequential A→B causal chain) \\
\textbf{Causal (model-based)} & DCM (fMRI) + Granger Causality (MEG) &
SN→DMN-ECN gating model must win Bayesian Model Selection; forward
Granger causality must exceed reverse \\
\end{longtable}
}

\textbf{Dual-track causal analysis.} The causal direction caveat
acknowledged in §4.4.3 remains: correlation between SN gating and
creative performance does not establish causation. While TMS causal
intervention is reserved for Phase 4, the present experiment
incorporates two complementary quasi-causal analyses operating at
different spatiotemporal scales.

\emph{fMRI track: Dynamic Causal Modeling (DCM).} Three competing models
of effective connectivity are specified over the DMN (mPFC, PCC), ECN
(dlPFC, PPC), and SN (dACC, aINS) nodes: \textbf{Model A} (SN gate)---SN
activity bilinearly modulates DMN$\leftrightarrow$ECN coupling strength during creative
task blocks; \textbf{Model B} (feedback)---DMN-ECN coupling changes
drive SN activity, with SN as a passive readout; \textbf{Model C}
(parallel)---all three networks respond independently to task demands
with no cross-network effective connectivity. Random-effects Bayesian
Model Selection (Stephan et al., 2009) adjudicates among models;
exceedance probability \(xp > 0.95\) for Model A constitutes strong
support for the SN gating hypothesis. The winning model's modulation
parameter is then correlated with \(\tau(X)\) across participants: the
prediction is a significant positive correlation (higher SN gate
strength → higher structural tension index).

\emph{MEG track: Conditional Granger Causality.} The millisecond
resolution of MEG enables directional analysis inaccessible to fMRI.
Conditional Granger causality
\(\text{GC}(\text{SN} \to \text{DMN} \mid \text{ECN})\) and
\(\text{GC}(\text{SN} \to \text{ECN} \mid \text{DMN})\) are computed on
source-reconstructed (LCMV beamformer) time series and compared against
the reverse directions
\(\text{GC}(\text{DMN} \to \text{SN} \mid \text{ECN})\) and
\(\text{GC}(\text{ECN} \to \text{SN} \mid \text{DMN})\). The prediction
is that forward Granger causality exceeds reverse in the overlap regime
but not in control conditions, with significance established by
phase-randomized surrogate testing (10,000 permutations). This analysis
directly complements the phase-amplitude coupling analysis in the
electrophysiological layer: SN theta phase should Granger-cause the
onset of gamma bursts in both DMN and ECN.

These quasi-causal analyses form a progressive evidential chain: if both
DCM (spatial, \textasciitilde seconds) and Granger causality (temporal,
\textasciitilde milliseconds) converge on SN→DMN-ECN directionality, the
case for TMS intervention in Phase 4 is substantially strengthened and
the specific stimulation target (dACC vs.~aINS) can be selected based on
the DCM node with the strongest modulation parameter.

\subsubsection{8.4 Experiments 3--4: Cross-Domain Isomorphism and Expert
Comparison}\label{experiments-34-cross-domain-isomorphism-and-expert-comparison}

\textbf{Experiment 3}: Using persistence diagrams as ``geometric
barcodes,'' compare topological signatures across visual-language,
auditory-motor, and spatial-conceptual domains. Cross-domain identical
Betti number features = support for the isomorphism conjecture. The
bottleneck distance \(W_\infty\) between persistence diagrams of
different modality pair conditions provides the quantitative decision
criterion.

\textbf{Experiment 4}: Expert-novice comparison---do experts trained in
one domain (e.g., visual-language) exhibit stronger Overlap Zone
signatures in domains where they have no systematic training? This
design tests whether cross-domain creative transfer operates via the
universal processing structure hypothesized herein. The prediction is
specific: expertise should transfer not at the level of content
(visual-language experts need not know anything about music) but at the
level of topological structure (the \(\beta_1\) persistence profile in
auditory-motor tasks should be elevated for visual-language experts
relative to novices). If transfer occurs at the content level but not
the structural level, the isomorphism conjecture is falsified.

\begin{center}\rule{0.5\linewidth}{0.5pt}\end{center}

\subsection{9. Discussion}\label{discussion}

\subsubsection{9.1 Limitations and Open
Problems}\label{limitations-and-open-problems}

Several fundamental limitations must be acknowledged. First, the fiber
bundle framework's constitutive objects (\(B\), \(G\), \(E_b\)) remain
at the level of candidate constructions rather than precisely determined
mathematical objects. The entire three-conjecture program is contingent
on advancing these definitions to proof-ready precision. Second, the
pathological mirror's empirical predictions rest on the assumption that
the creativity-psychopathology spectrum can be adequately captured by a
two-dimensional parameter space (coupling intensity × regulatory
capacity); the true dimensionality may be higher, with additional
factors such as attentional modulation, developmental history, and
pharmacological state playing independent roles.

Third, the computational architecture's reliance on persistent homology
as a differentiable training signal involves the \(O(n^3)\) bottleneck
and the challenge of gradient propagation through discrete topological
computations. While the differentiable topological layer of Hofer et
al.~(2019) provides a principled approach and the scalability roadmap in
§6.2 outlines a staged migration path (Witness Complex, DTM filtration),
scaling to production-level model sizes with millions of parameters
remains unvalidated. Fourth, the ANALOGY-MM benchmark, while
theoretically motivated, has not yet undergone psychometric
validation---the item difficulty, discriminability, and cultural
fairness of the cross-modal analogy tasks require empirical assessment.
The forced-choice protocol introduced in §7.3 addresses the scalability
of ETR computation but introduces its own limitations: forced-choice
formats may suppress overflow errors that are themselves diagnostic of
contact topology, a concern partially mitigated by the open-generation
calibration subset.

\subsubsection{9.2 Long-Term Vision: From Architecture to
Application}\label{long-term-vision-from-architecture-to-application}

If the synthetic data PoC and early behavioral experiments yield
positive results, the long-term implications extend well beyond
multimodal AI architecture. Protein structure prediction, for instance,
currently operates within a separable representational paradigm
(sequence → structure mapping). The Overlap Zone framework suggests that
a class of protein \emph{function} prediction tasks---those involving
allosteric regulation, intrinsically disordered regions, and
context-dependent conformational dynamics---may require non-separable
representations of sequence, structure, and dynamics. This is a testable
prediction that bridges the current framework to one of the most
consequential domains in computational biology.

Similarly, the framework has implications for scientific discovery
systems that must integrate heterogeneous data sources (genomic,
proteomic, metabolomic, clinical) in ways that go beyond concatenation
or late fusion. If the structural insight is correct---that certain
scientific questions require representations in which the information
from different sources is constitutively inseparable---then the
topological regularization approach may find application far beyond the
creative cognition domain that motivates it. Drug discovery, for
instance, increasingly requires simultaneous modeling of molecular
structure, binding dynamics, cellular context, and patient-level
variation---a multi-scale integration problem where the relevant
information is distributed across representations that cannot be
meaningfully decomposed into independent modality-specific features.
Materials science presents analogous challenges, where the relationship
between atomic structure, mesoscale organization, and macroscopic
properties is constitutively non-separable in ways that current
multi-scale simulation approaches approximate but cannot fully capture.

\subsubsection{9.3 Relation to Broader Debates in
AI}\label{relation-to-broader-debates-in-ai}

The contact topology diagnosis contributes to the ongoing debate about
the nature of understanding in large language models. The Overlap Zone
framework provides a precise structural claim: systems operating
exclusively in the contact regime can exhibit impressive performance on
tasks decomposable into propositional subtasks, but will systematically
fail on tasks requiring the simultaneous, inseparable co-presence of
propositional and non-propositional processing. This is not a claim
about ``consciousness'' or ``understanding'' in the philosophical
sense---it is a claim about \emph{representational topology} with
specific, testable computational predictions.

\subsubsection{9.4 Ethical Considerations}\label{ethical-considerations}

The pathological mirror construct raises ethical considerations that
must be addressed proactively. Operationalizing the
creativity-psychopathology spectrum computationally---through structural
tension indices and topological signatures---risks reductive treatment
of clinical populations. We emphasize that the framework proposes a
\emph{structural} analogy (shared parameter space, different regions)
rather than a \emph{clinical} claim (creative people are pathological,
or pathological people are creative). All neuroimaging and behavioral
experiments involving clinical or at-risk populations will require
institutional ethics board approval and must adhere to the highest
standards of informed consent and data protection.

\begin{center}\rule{0.5\linewidth}{0.5pt}\end{center}

\subsection{10. Conclusion}\label{conclusion}

This paper identifies a structural limitation in current multimodal AI
that is topological rather than parametric: the systematic suppression
of non-separable cross-modal representations through training objectives
and inductive biases. We term this missing representational mechanism
\emph{overlap topology} and propose to formalize it via connection
theory on fiber bundles over the modality configuration space. By
introducing the topological opposition between overlap isomorphism and
superimposition collapse, the theory acquires a precise falsification
condition: if highly creative individuals and those with high schizotypy
exhibit topologically indistinguishable signature markers in neural
representation space---i.e., \(\tau(X_{\text{creative}})\) and
\(\tau(X_{\text{pathological}})\) are statistically
indistinguishable---the theory's core distinction collapses.

The computational proposal---Neural ODEs with topological
regularization---directly responds to this diagnosis: not building an
entirely new architecture, but designing training objectives that force
existing architectures to maintain non-separable representations. The
synthetic data PoC (executable within weeks) provides a low-cost entry
point; the ANALOGY-MM benchmark provides an immediately deployable
architectural diagnostic. The phased experimental agenda ensures that
each escalation in cost and complexity is contingent on prior positive
results, with explicit falsification criteria at every gate.

What this paper truly seeks is not a new architecture for its own sake,
but an empirical answer to a precise question: does there exist a class
of cognitive and computational operations whose optimal performance
essentially requires structurally inseparable representations? If yes,
the implications for AI architecture are profound---and the pathological
mirror opens new connections between computational psychiatry and
multimodal AI safety. If no, the program exits cleanly at minimal cost,
while contributing a precise topological vocabulary for analyzing
multimodal integration strategies along the way.

\begin{center}\rule{0.5\linewidth}{0.5pt}\end{center}

\subsection{Appendix A: Precise Definitions of Core
Terms}\label{appendix-a-precise-definitions-of-core-terms}

\begin{enumerate}
\def\labelenumi{\arabic{enumi}.}
\item
  \textbf{Contact Topology}: A representational regime in which
  information from different modalities meets at low-dimensional
  boundaries of measure zero. The common geometric prior of current
  multimodal AI architectures.
\item
  \textbf{Overlap Topology}: A representational regime in which elements
  from different modalities are structurally inseparable within a shared
  region of non-zero interior, mutually constituting each other's
  semantic content.
\item
  \textbf{Overlap Zone}: A cognitive processing region or
  representational state in which propositional and non-propositional
  content are simultaneously co-present and structurally inseparable.
\item
  \textbf{Superimposition Collapse}: The pathological mirror of the
  Overlap Zone---dissolution of the structural boundary between
  propositional and non-propositional dimensions, loss of
  transversality, and destruction of generative tension.
\item
  \textbf{Transversality}: The condition under which two subvarieties
  intersect while preserving each other's local structural integrity.
  Overlap corresponds to transversal coupling; collapse corresponds to
  loss of transversality.
\item
  \textbf{Creative Transformation (\emph{Chuànghuà})}: The
  meaning-emergence event co-generated by propositional and
  non-propositional dimensions within the Overlap Zone.
\item
  \textbf{Xiàng (Operative Schema)}: An operative schema from the
  Chinese epistemological tradition---a dynamic structure that
  simultaneously organizes perceptual, conceptual, and procedural
  knowledge. Identified as a stable section within the fiber bundle
  framework.
\item
  \textbf{Inseparability Signature}: The core experimental prediction of
  the Overlap Zone---selective perturbation of one dimension produces
  semantically related synchronous covariation in the other dimension
  (not merely resource-competitive impairment).
\item
  \textbf{Non-Separability Entropy (NS)}:
  \(\text{NS}(z) = -\sum_i p_i \log p_i\), the Shannon entropy of
  squared Schmidt decomposition coefficients. NS \(= 0\) iff \(z\) is
  fully separable.
\item
  \textbf{Structural Tension Index (\(\tau(X)\))}: \(\beta_1\) total
  persistence / \(\beta_0\) total persistence, an operationalized
  indicator extracted from persistent homology.
\item
  \textbf{Universal Overlap Zone Operator (UOO)}: An operator satisfying
  three conditions: (i) maintaining inseparability (NS \(\geq \kappa\));
  (ii) bidirectional semantic coupling; (iii) cross-domain structural
  invariance.
\item
  \textbf{Yang-Mills Three-Regime Model}: Three dynamical regimes
  classified by \(\|F_\nabla\|^2\): I (contact, no connection), II
  (overlap, \(0 < \|F_\nabla\|^2 < C\)), III (collapse,
  \(\|F_\nabla\|^2 \geq C\)).
\item
  \textbf{Dual-Threshold Model}: Three-interval partition in NS-space:
  \(\text{NS} < \kappa^*\), \(\kappa^* \leq \text{NS} < \kappa^{**}\),
  \(\text{NS} \geq \kappa^{**}\).
\item
  \textbf{Shared Vulnerability Model}: Carson's (2011) framework:
  creativity and psychopathology share cognitive features; protective
  factors determine the outcome direction. Reinterpreted as a 2D
  parameter space (coupling intensity × regulatory capacity).
\item
  \textbf{ANALOGY-MM}: Cross-modal analogy evaluation benchmark. Core
  metric is ETR (error type ratio), detecting the ``content acquired but
  structural mapping failed'' error bias characteristic of contact
  topology. Scalable evaluation via 2×2 factorial forced-choice protocol
  (§7.3).
\item
  \textbf{Dynamic Causal Modeling (DCM)}: A Bayesian framework for
  estimating effective (directed) connectivity among brain regions from
  fMRI data. Used in §8.3 to adjudicate among competing models of
  SN→DMN-ECN gating.
\item
  \textbf{Granger Causality}: A statistical test for directed temporal
  dependencies in time series. Applied to MEG source-reconstructed
  signals to test SN→DMN-ECN directionality at millisecond resolution
  (§8.3).
\item
  \textbf{Witness Complex}: A computationally efficient approximation to
  the Vietoris-Rips complex, constructed from a subset of landmark
  points. Enables scaling of persistent homology computation from
  \(O(n^3)\) to \(O(m^3)\) with \(m \ll n\) (§6.2).
\item
  \textbf{Cruciform Structure (shizi jiegou)}: The philosophical framework
  in which \emph{xiàng} occupies the intersection of two axes:
  \emph{dào}$\leftrightarrow$\emph{qì} (vertical, metaphysical$\leftrightarrow$physical) and
  saying$\leftrightarrow$showing (horizontal, propositional$\leftrightarrow$presentational). §3.3
  develops this structure; §5.8 provides its fiber bundle mapping.
\item
  \textbf{\emph{Chuànghuà} (chuanghua, Creative Transformation)}: The
  spontaneous meaning-emergence event at the saying/showing
  intersection. Computationally corresponds to the Neural ODE trajectory
  in which non-trivial \(\beta_1\) topology dynamically emerges.
\item
  \textbf{\emph{Huàcái} (huacai, Institutional Cutting)}: The
  systematization of \emph{chuànghuà} into repeatable, transmissible
  form. Computationally corresponds to the frozen parameters
  \(\theta^*\) encoding the learned connection.
\item
  \textbf{META-TOP (Metaphor-Topology)}: Three-tier benchmark extending
  ANALOGY-MM. Tests progressively deeper cross-modal topological
  understanding, from structural mapping (L1) through dynamical pattern
  recognition (L2) to cross-civilizational topological isomorphism (L3).
  Core metric: TSAS (Topological Structural Alignment Score).
\end{enumerate}

\begin{center}\rule{0.5\linewidth}{0.5pt}\end{center}

\subsection{Appendix B: Synthetic Validation Pseudocode}\label{appendix-b-pseudocode}

The following annotated pseudocode outlines the synthetic data proof of concept described in \S6.3. All components use standard PyTorch modules; the only non-standard dependency is a differentiable persistent homology layer (Hofer et al., 2017).

\begin{small}
\begin{verbatim}
# === UOO Toy PoC: Neural ODE + Topological Regularization ===
# Reference implementation for the three-condition experiment (§6.3)

import torch
import torch.nn as nn
from torchdiffeq import odeint_adjoint as odeint  # Chen et al. 2018
from topological_layer import RipsLayer            # Hofer et al. 2017

# --- Phase 1: Bilinear Entanglement (Eq. 1) ---
class BilinearEntangle(nn.Module):
    """Maps (x, y) -> z(0) via learnable bilinear kernel.

    Mathematical correspondence: implements the tensor product projection
    W_entangle(x tensor y) from S6.2. nn.Bilinear performs implicit Tucker
    decomposition (rank D); set D = d1*d2 for exact tensor product,
    or D < d1*d2 as compressed approximation (recommended for d1,d2 > 32).
    """
    def __init__(self, d1, d2, D):
        super().__init__()
        self.W = nn.Bilinear(d1, d2, D)
    def forward(self, x, y):
        return self.W(x, y)  # z(0) in R^D

# --- Phase 2: Neural ODE as Fiber Bundle Solver (Eq. 2) ---
class ODEFunc(nn.Module):
    """Parameterizes dz/dt = f_theta(z, t), the connection 1-form.

    The time variable t is concatenated to z, enabling the network to
    learn phase-dependent dynamics (fusion -> oscillation -> stabilization).
    SiLU activation chosen for smoothness (C^inf), ensuring the ODE
    satisfies Lipschitz conditions for existence/uniqueness.
    """
    def __init__(self, D, hidden=256):
        super().__init__()
        self.net = nn.Sequential(
            nn.Linear(D + 1, hidden), nn.SiLU(),
            nn.Linear(hidden, hidden), nn.SiLU(),
            nn.Linear(hidden, D))
    def forward(self, t, z):
        return self.net(torch.cat([z, t.expand(z.shape[0], 1)], -1))

# --- Phase 3: Topological Regularization (Eq. 3) ---
def topo_loss(z, rips_layer, lam=1.0, eps_min=1e-4):
    """Compute L_topo from persistence diagram of z(T).

    Penalizes beta_0 (fragmentation) and rewards beta_1 (cross-modal
    loops). eps_min threshold filters near-degenerate persistence
    features to avoid gradient instability (see §6.2 gradient analysis).
    """
    dgm = rips_layer(z)
    b0 = dgm[0]  # beta_0 lifetimes
    b1 = dgm[1]  # beta_1 lifetimes
    # Filter near-degenerate features
    b0 = b0[b0 > eps_min]
    b1 = b1[b1 > eps_min]
    L = (b0**2).sum() - lam * (b1**2).sum()
    return L

# --- Full UOO Forward Pass ---
def uoo_forward(x, y, entangle, ode_fn, rips, alpha=0.1):
    """Complete UOO pipeline: entangle -> evolve -> regularize.

    Set alpha=0 for the ablation condition (c) in §6.3.
    """
    z0 = entangle(x, y)                        # Phase 1
    t  = torch.linspace(0, 1, 20)
    zT = odeint(ode_fn, z0, t)[-1]             # Phase 2
    L_task = task_loss(zT, target)
    L_topo = topo_loss(zT, rips)                # Phase 3
    L_total = L_task + alpha * L_topo           # Eq. (4)
    return L_total, zT

# --- Evaluation Metrics ---
def structural_tension(z, rips_layer):
    """tau(X) = beta_1 total persistence / beta_0 total persistence.

    The primary topological diagnostic: tau > 1 indicates cross-modal
    loops dominate over fragmentation (overlap regime signature).
    """
    dgm = rips_layer(z)
    tau = dgm[1].sum() / (dgm[0].sum() + 1e-8)
    return tau

def non_separability_entropy(z, d1, d2):
    """NS(z) via Schmidt decomposition of cross-modal representation.

    Reshape z to (d1, d2) matrix, compute SVD, derive Shannon entropy
    of squared singular values. NS = 0 iff z is separable (rank-1).
    Requires D = d1*d2; use projection if D != d1*d2.
    """
    U, S, V = torch.linalg.svd(z.view(-1, d1, d2))
    p = (S**2) / (S**2).sum(-1, keepdim=True)
    NS = -(p * p.log()).sum(-1).mean()
    return NS
\end{verbatim}
\end{small}

\noindent\textbf{Experimental protocol.} (1)~Generate two synthetic modalities: $\mathbf{x} \in \mathbb{R}^{64}$ (``visual'': geometric patterns) and $\mathbf{y} \in \mathbb{R}^{64}$ (``verbal'': symbolic sequences) with shared latent structure. (2)~Train three conditions: (a)~UOO (Neural ODE + $\mathcal{L}_{\text{topo}}$, $\alpha = 0.1$), (b)~Neural ODE alone ($\alpha = 0$), and (c)~CLIP-style contrastive baseline, all matched in parameters ($\sim$500K). (3)~Monitor $\tau(X)$, NS, and $\beta_1$ profiles throughout training (log every 50 steps). (4)~Test zero-shot transfer on novel modality pairs. \textbf{Termination}: if condition~(a) shows no statistically significant advantage in $\tau(X)$ or zero-shot transfer after convergence (200 epochs with no improvement), the program terminates.

\begin{center}\rule{0.5\linewidth}{0.5pt}\end{center}

\subsection{References}\label{references}
\RaggedRight
\small

Acar, S., \& Sen, S. (2013). A multilevel meta-analysis of the
relationship between creativity and schizotypy. \emph{Psychology of
Aesthetics, Creativity, and the Arts}, 7(3), 214--228.

Adams, R. A., Stephan, K. E., Brown, H. R., Frith, C. D., \& Friston, K.
J. (2013). The computational anatomy of psychosis. \emph{Frontiers in
Psychiatry}, 4, 47.

Alayrac, J.-B., Donahue, J., Luc, P., Miech, A., Barr, I., Hasson, Y.,
\ldots, \& Simonyan, K. (2022). Flamingo: A visual language model for
few-shot learning. \emph{NeurIPS}, 35.

Alon, U., \& Yahav, E. (2021). On the bottleneck of graph neural
networks and its practical implications. \emph{Proceedings of ICLR}.

Anai, H., Chazal, F., Glisse, M., Ike, Y., Inakoshi, H., Tinarrage, R.,
\& Umeda, Y. (2020). DTM-based filtrations. In \emph{Topological Data
Analysis} (pp.~33--66). Springer.

Anticevic, A., Cole, M. W., Murray, J. D., Corlett, P. R., Wang, X.-J.,
\& Krystal, J. H. (2012). The role of default network deactivation in
cognition and disease. \emph{Trends in Cognitive Sciences}, 16(12),
584--592.

Bauer, U. (2021). Ripser: Efficient computation of Vietoris-Rips
persistence barcodes. \emph{Journal of Applied and Computational
Topology}, 5, 391--423.

Beaty, R. E., Kenett, Y. N., Christensen, A. P., Rosenberg, M. D.,
Benedek, M., Chen, Q., Fink, A., Qiu, J., Kwapil, T. R., Kane, M. J., \&
Silvia, P. J. (2018). Robust prediction of individual creative ability
from brain functional connectivity. \emph{PNAS}, 115(5), 1087--1092.

Beaty, R. E., Seli, P., \& Schacter, D. L. (2019). Network neuroscience
of creative cognition. \emph{Current Opinion in Behavioral Sciences},
27, 22--30.

Belting, H. (2011). \emph{An Anthropology of Images: Picture, Medium,
Body}. Princeton University Press.

Benedek, M., \& Fink, A. (2019). Toward a neurocognitive framework of
creative cognition. \emph{Current Opinion in Behavioral Sciences}, 27,
116--122.

Boehm, G. (1994). Die Wiederkehr der Bilder. In G. Boehm (Ed.),
\emph{Was ist ein Bild?} (pp.~11--38). Fink.

Bronstein, M. M., Bruna, J., Cohen, T., \& Veličković, P. (2021).
Geometric Deep Learning: Grids, Groups, Graphs, Geodesics, and Gauges.
\emph{arXiv:2104.13478}.

Carrière, M., Chazal, F., Ike, Y., Lacombe, T., Royer, M., \& Umeda, Y. (2020). PersLay: A neural network
layer for persistence diagrams. \emph{Proceedings of AISTATS}.

Carson, S. H. (2011). Creativity and psychopathology: A shared
vulnerability model. \emph{Canadian Journal of Psychiatry}, 56(3),
144--153.

Cohen, T. S., Weiler, M., Kicanaoglu, B., \& Welling, M. (2019). Gauge
equivariant convolutional networks and the icosahedral CNN. \emph{Proceedings of ICML}.

de Silva, V., \& Carlsson, G. (2004). Topological estimation using
witness complexes. In \emph{Symposium on Point-Based Graphics}
(pp.~157--166). Eurographics.

Di Giovanni, F., Rowbottom, J., Chamberlain, B. P., Dong, X., \&
Bronstein, M. M. (2023). Understanding convolution on graphs via
energies. \emph{Transactions on Machine Learning Research}.

Edelsbrunner, H., \& Harer, J. (2010). \emph{Computational Topology: An
Introduction}. AMS.

Eells, J., \& Sampson, J. H. (1964). Harmonic mappings of Riemannian
manifolds. \emph{American Journal of Mathematics}, 86(1), 109--160.

Evans, R. (1997). \emph{Translations from Drawing to Building and Other
Essays}. MIT Press.

Gemini Team. (2024). Gemini: A family of highly capable multimodal
models. \emph{arXiv:2312.11805}.

Gentner, D. (1983). Structure-mapping: A theoretical framework for
analogy. \emph{Cognitive Science}, 7(2), 155--170.

Gilmer, J., Schoenholz, S. S., Riley, P. F., Vinyals, O., \& Dahl, G. E.
(2017). Neural message passing for quantum chemistry. \emph{Proceedings
of ICML}.

Hofer, C., Kwitt, R., Niethammer, M., \& Uhl, A. (2017). Deep learning
with topological signatures. \emph{NeurIPS}, 30.

Horodecki, R., Horodecki, P., Horodecki, M., \& Horodecki, K. (2009).
Quantum entanglement. \emph{Reviews of Modern Physics}, 81(2), 865--942.

Jia, C., Yang, Y., Xia, Y., Chen, Y.-T., Parekh, Z., Pham, H., \ldots,
\& Duerig, T. (2021). Scaling up visual and vision-language
representation learning with noisy text supervision. \emph{Proceedings
of ICML}.

Jost, J. (2017). \emph{Riemannian Geometry and Geometric Analysis} (7th
ed.). Springer.

Jullien, F. (1995). \emph{The Propensity of Things: Toward a History of
Efficacy in China}. Zone Books.

Kyaga, S., Lichtenstein, P., Boman, M., Hultman, C., Långström, N., \&
Landén, M. (2011). Creativity and mental disorder. \emph{British Journal
of Psychiatry}, 199(5), 373--379.

Lake, B. M., Salakhutdinov, R., \& Tenenbaum, J. B. (2015). Human-level
concept learning through probabilistic program induction.
\emph{Science}, 350(6266), 1332--1338.

Menon, V. (2011). Large-scale brain networks and psychopathology: A
unifying triple network model. \emph{Trends in Cognitive Sciences},
15(10), 483--506.

Mitchell, W. J. T. (1994). \emph{Picture Theory: Essays on Verbal and
Visual Representation}. University of Chicago Press.

Moor, M., Horn, M., Rieck, B., \& Borgwardt, K. (2020). Topological
Autoencoders. \emph{Proceedings of ICML}.

Ngiam, J., Khosla, A., Kim, M., Nam, J., Lee, H., \& Ng, A. Y. (2011).
Multimodal deep learning. \emph{Proceedings of ICML}.

OpenAI. (2023). GPT-4 Technical Report. \emph{arXiv:2303.08774}.

Palaniyappan, L., \& Liddle, P. F. (2012). Does the salience network
play a cardinal role in psychosis? \emph{Journal of Psychiatry \&
Neuroscience}, 37(1), 17--27.

Petri, G., Expert, P., Turkheimer, F., Carhart-Harris, R., Nutt, D.,
Hellyer, P. J., \& Vaccarino, F. (2014). Homological scaffolds of brain
functional networks. \emph{Journal of the Royal Society Interface},
11(101), 20140873.

Radford, A., Kim, J. W., Hallacy, C., Ramesh, A., Goh, G., Agarwal, S.,
\ldots, \& Sutskever, I. (2021). Learning transferable visual models
from natural language supervision. \emph{Proceedings of ICML}.

Ramesh, A., Dhariwal, P., Nichol, A., Chu, C., \& Chen, M. (2022).
Hierarchical text-conditional image generation with CLIP latents.
\emph{arXiv:2204.06125}.

Saggar, M., Sporns, O., Gonzalez-Castillo, J., Bandettini, P. A.,
Carlsson, G., Glover, G., \& Reiss, A. L. (2018). Towards a new approach
to reveal dynamical organization of the brain using topological data
analysis. \emph{Nature Communications}, 9, 1399.

Saharia, C., Chan, W., Saxena, S., Li, L., Whang, J., Denton, E. L.,
\ldots, \& Norouzi, M. (2022). Photorealistic text-to-image diffusion
models with deep language understanding. \emph{NeurIPS}, 35.

Stephan, K. E., Penny, W. D., Daunizeau, J., Moran, R. J., \& Friston,
K. J. (2009). Bayesian model selection for group studies.
\emph{NeuroImage}, 46(4), 1004--1017.

Stolz, B. J., Harrington, H. A., \& Porter, M. A. (2017). Persistent
homology of time-dependent functional networks. \emph{Chaos}, 27(4),
047410.

Tan, X. (2008). \emph{Illustrating Architectonics: Pictorial Philosophy
in Architectural Perspectives} {[}Doctoral dissertation{]}. South China
University of Technology.

Tan, X. (2026). Jiaodiequ Tonggou Xing yu Tongyong Jiaodiequ Suanzi(quanwen xiuding ban){[}Overlap
Zone Isomorphism and the UOO{]}. Unpublished companion manuscript.

Tan, X. (2026b). Jiaodiequ Tonggou Xing {[}Overlap Zone Isomorphism: On the
Creative-Transformative Zone of Image{]}. Unpublished companion
monograph.

Taylor, C. L. (2017). Creativity and mood disorder: A systematic review
and meta-analysis. \emph{Perspectives on Psychological Science}, 12(6),
1040--1076.

Topping, J., Di Giovanni, F., Chamberlain, B. P., Dong, X., \&
Bronstein, M. M. (2022). Understanding over-squashing and bottlenecks on
graphs via curvature. \emph{Proceedings of ICLR}.

Weiler, M., \& Cesa, G. (2019). General E(2)-equivariant steerable CNNs.
\emph{NeurIPS}, 32.

Whitfield-Gabrieli, S., \& Ford, J. M. (2012). Default mode network
activity and connectivity in psychopathology. \emph{Annual Review of
Clinical Psychology}, 8, 49--76.

Wittgenstein, L. (1922). \emph{Tractatus Logico-Philosophicus}. Kegan
Paul.

\begin{center}\rule{0.5\linewidth}{0.5pt}\end{center}

\emph{This paper is a working draft. The mathematical sections (§5)
require elevation to the precision needed for mathematicians to initiate
formal work; the computational sections (§6) await implementation and
validation; the experimental sections (§8) require ethical approval and
pilot testing. We warmly welcome collaborators from all three pillar
domains.}

\emph{Correspondence: Guangzhou Academy of Fine Arts}

\end{document}